\documentclass[conference]{IEEEtran}
\IEEEoverridecommandlockouts
\usepackage{cite}
\usepackage{amsmath,amssymb,amsfonts}
\usepackage{algorithmic}
\usepackage{graphicx}
\usepackage{textcomp}
\usepackage{xcolor}
\usepackage{xspace}
\usepackage{enumitem}
\usepackage{subfigure}
\usepackage{booktabs}
\usepackage{multirow}
\usepackage{colortbl}
\definecolor{LightCyan}{rgb}{0.8,0.9,1}
\usepackage{hyperref}

\usepackage{orcidlink}

\def\BibTeX{{\rm B\kern-.05em{\sc i\kern-.025em b}\kern-.08em
    T\kern-.1667em\lower.7ex\hbox{E}\kern-.125emX}}

\makeatletter
\DeclareRobustCommand\bmvaOneDot{\futurelet\@let@token\bmv@onedotaux}
\def\bmv@onedotaux{\ifx\@let@token.\else.\null\fi\xspace}
\makeatother
\def\eg{\emph{e.g}\bmvaOneDot}

\def\etal{\emph{et al}\bmvaOneDot}
\def\etc{\emph{etc}\bmvaOneDot}

\usepackage{amsmath,amsfonts,bm}

\newcommand{\idx}[1]{\mathcal{I}_{#1}}

\newcommand{\mbr}[1]{\mathbb{R}^{#1}}

\def\eqref#1{(\ref{#1})}
\def\1{\bm{1}}

\def\vx{{\bm{x}}}

\def\mA{{\bm{A}}}

\def\mD{{\bm{D}}}

\def\mF{{\bm{F}}}

\def\mV{{\bm{V}}}

\def\mX{{\bm{X}}}

\DeclareMathAlphabet{\mathsfit}{\encodingdefault}{\sfdefault}{m}{sl}
\SetMathAlphabet{\mathsfit}{bold}{\encodingdefault}{\sfdefault}{bx}{n}
\newcommand{\tens}[1]{\bm{\mathsfit{#1}}}
\def\tA{{\tens{A}}}

\def\tD{{\tens{D}}}

\def\tF{{\tens{F}}}

\def\tH{{\tens{H}}}

\def\tV{{\tens{V}}}

\def\tX{{\tens{X}}}

\begin{document}

\title{TrackNetV4: Enhancing Fast Sports Object Tracking with Motion Attention Maps \\
% Enhancing Tracking of Fast Sports Objects with Motion Attention Maps\\
}

\author{\IEEEauthorblockN{Arjun Raj\orcidlink{0009-0003-8717-6307}}
\IEEEauthorblockA{\textit{School of Computing} \\
\textit{Australian National University}\\
Canberra, Australia \\
u7526852@anu.edu.au
% email address or ORCID
}
\and
\IEEEauthorblockN{Lei Wang\thanks{* Corresponding author. Our project website is \href{https://time.anu.edu.au/paper-sites/tracknet-v4}{here}.} \textsuperscript{$\!\!*$}\orcidlink{0000-0002-8600-7099}}
\IEEEauthorblockA{\textit{School of Computing} \\
\textit{Australian National University}\\
Canberra, Australia \\
lei.w@anu.edu.au 
% email address or ORCID
}
\and
\IEEEauthorblockN{Tom Gedeon\orcidlink{0000-0001-8356-4909}}
\IEEEauthorblockA{\textit{School of Elec Eng, Comp \& Math Sci} \\
\textit{Curtin University}\\
Perth, Australia \\
tom.gedeon@curtin.edu.au
% email address or ORCID
}
}

\maketitle

\begin{abstract}

Accurately detecting and tracking high-speed, small objects, such as balls in sports videos, is challenging due to factors like motion blur and occlusion. Although recent deep learning frameworks like TrackNetV1, V2, and V3 have advanced tennis ball and shuttlecock tracking, they often struggle in scenarios with partial occlusion or low visibility. This is primarily because these models rely heavily on visual features without explicitly incorporating motion information, which is crucial for precise tracking and trajectory prediction. In this paper, we introduce an enhancement to the TrackNet family by fusing high-level visual features with learnable motion attention maps through a motion-aware fusion mechanism, effectively emphasizing the moving ball's location and improving tracking performance. Our approach leverages frame differencing maps, modulated by a motion prompt layer, to highlight key motion regions over time. Experimental results on the tennis ball and shuttlecock datasets show that our method enhances the tracking performance of both TrackNetV2 and V3. We refer to our lightweight, plug-and-play solution, built on top of the existing TrackNet, as TrackNetV4.

\end{abstract}

\begin{IEEEkeywords}
tracking, motion attention, fusion
\end{IEEEkeywords}

\section{Introduction}

Ball trajectory data is a crucial element in sports analysis and athlete training. However, accurately detecting and tracking high-speed, small balls in sports competition videos presents significant challenges. The primary difficulties arise from the fact that balls in broadcast videos often appear blurry, tiny, or obscured by afterimages. Additionally, they may become invisible due to occlusion, extreme visual indistinctness, or simply flying out of the camera's field of view.

With recent advances in deep learning, TrackNetV1 is introduced in~\cite{huang2019tracknet} to track tennis balls and shuttlecocks in broadcast match videos. This heatmap-based tracking framework, built on a VGG-16 feature extraction network~\cite{simonyan2014very} and an upsampling network~\cite{noh2015learning}, takes multiple consecutive frames as input, leveraging the ball's trajectory\footnote{The ball's trajectory is determined by a sequence of video frames, yet this process does not explicitly incorporate any form of motion cues.} for improved detection. While TrackNetV1 achieves superior tracking performance compared to conventional methods, its processing speed is insufficient for real-time sports analysis, and its network design consumes substantial GPU memory.

\begin{figure}[t]%htbp % left bottom right top
\centering%%%%
\begin{tabular}[t]{ccc}
\subfigure[RGB frame]
{\label{fig:rgb}\includegraphics[trim=0cm 0cm 0cm 0cm, clip=true, width=0.32\linewidth]{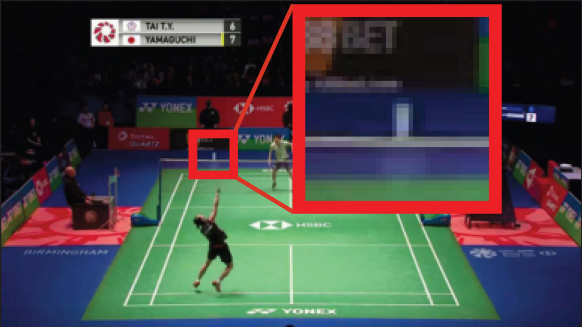}}
\subfigure[Attention map]
{\label{fig:attn}\includegraphics[trim=0cm 0cm 0cm 0cm, clip=true, width=0.32\linewidth]{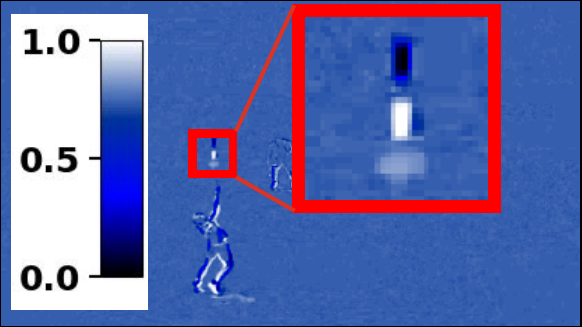}}
\subfigure[Motion prompt]
{\label{fig:prompt}\includegraphics[trim=0cm 0cm 0cm 0cm, clip=true, width=0.32\linewidth]{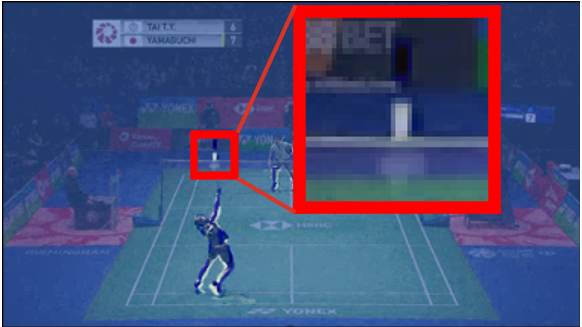}}
\end{tabular}
\caption{A visual comparison is presented between (a) the original video frame, (b) the learned motion attention map, and (c) the motion-prompted frame. 
Tracking shuttlecocks is challenging due to their small size and tendency to blend into the background. To address this, we use a motion prompt layer~\cite{chen2024motion} to generate motion attention maps that highlight the shuttlecock's location. We also create a motion-prompted frame by performing element-wise multiplication between the motion attention map and the original video frame, showing how motion features enhance visual representation. For better visualization, the shuttlecocks in these frames are zoomed in on the right. % Best viewed in color.
} 
\label{fig:attn-prompt}
\end{figure}

TrackNetV2, presented in~\cite{Sun2020TrackNetV2ES}, offers improvements over TrackNetV1 by: (i) increasing processing speed through reduced input size (\eg, from 640$\times$360 to 512$\times$288) and a redesigned multiple-in, multiple-out architecture (\eg, from 3 in 1 out to 3 in 3 out), (ii) enhancing tracking accuracy by introducing and training on a comprehensive badminton match video dataset, and (iii) lowering GPU memory usage by replacing a pixel-wise one-hot encoding 3D array with a real-valued 2D array. Additionally, TrackNetV2 introduces a weighted cross-entropy loss function to focus on ball movements more effectively, and uses skip connections to preserve tiny object information, preventing the degradation of small object features in the network. Experiments show that TrackNetV2 significantly improves both trajectory prediction and real-time processing speed for badminton tracking.
Similar to TrackNetV1, TrackNetV2 still does not explicitly consider motion. Its tracking and prediction performance are enhanced through a few skip connections that fuse low- and high-level feature maps extracted solely from video frames.

% Spanning figure across both columns
\begin{figure*}[htbp]
    \centering
    \includegraphics[width=\textwidth]{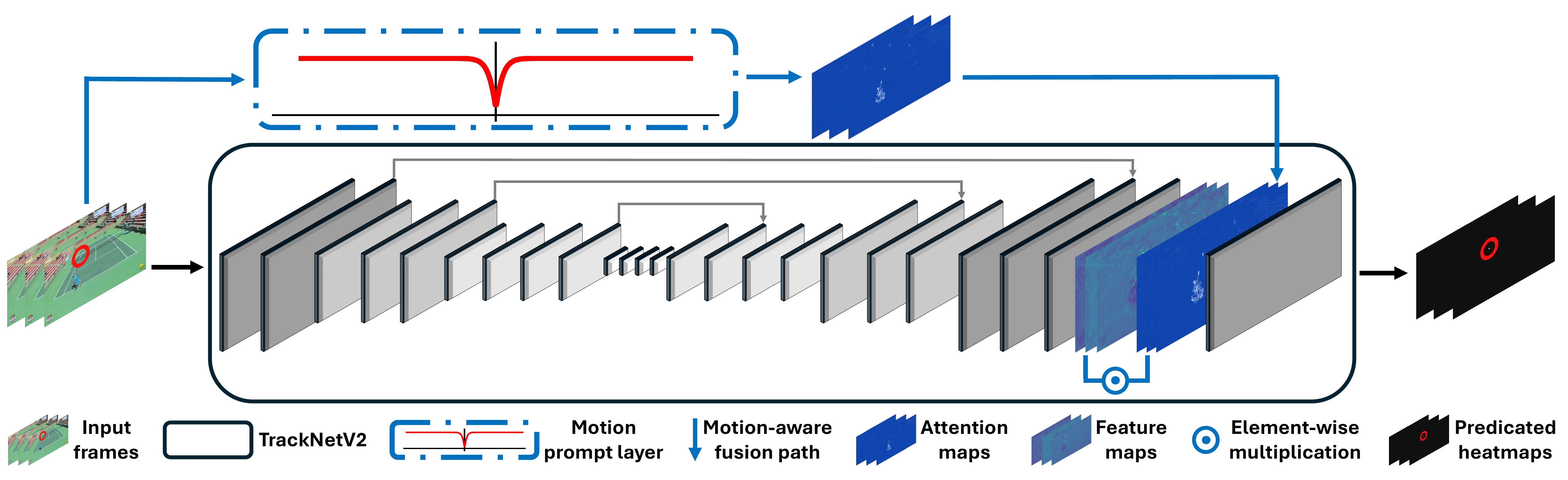}
    \caption{We propose using learnable motion attention maps to enhance the tracking of high-speed, small objects in video frames. While demonstrated with TrackNetV2, our approach can be seamlessly integrated into any heatmap-based detection and tracking framework.
    Our method uses a motion prompt layer~\cite{chen2024motion} on frame differencing maps (using absolute values to capture both positive and negative pixel intensity changes, thereby reducing missed detections) to generate motion attention maps that highlight key motion regions, such as balls. These maps are then fused with high-level visual features before the heatmap output layer through element-wise multiplication, followed by concatenation. The tracking framework that features our motion-aware fusion is named TrackNetV4.}
    \label{fig:main}
\end{figure*}

Recently, TrackNetV3~\cite{10.1145/3595916.3626370} further improved ball tracking accuracy and trajectory completeness, even when the ball is temporarily obstructed. Compared to TrackNetV2, TrackNetV3 incorporates estimated background information as auxiliary data to better locate the ball and introduces a trajectory rectification module that interpolates missing ball coordinates using image inpainting techniques. TrackNetV3 surpasses baselines like TrackNetV2 and YOLOv7~\cite{wang2023yolov7} by implicitly using motion dynamics, for example, through the subtraction between the original video frame and the estimated background image. However, it still struggles when the ball is barely visible, relying heavily on temporal dynamics from consecutive frames, which often contain noise and irrelevant motion information.

Ball tracking and prediction in sports analysis are highly dependent on effective temporal dynamics extraction, yet even the state-of-the-art TrackNetV3 does not explicitly leverage motion information. Inspired by the recent success of prompts in computer vision tasks and the use of motion as learnable prompts~\cite{chen2024motion} to selectively focus on relevant motions while suppressing unwanted noise, we propose a simple yet innovative module that uses learnable motion attention maps to enhance the tracking.
Specifically, our approach: (i) selectively highlights motion regions of interest in video frames, and (ii) introduces motion-aware fusion mechanism to preserve the motion regions of high-speed, small objects.

Our contributions are summarized as follows:
\renewcommand{\labelenumi}{\roman{enumi}.}
\begin{enumerate}[leftmargin=0.4cm]
    \item We are the first to incorporate learnable motion attention maps into the tracking framework, enabling it to focus on movements crucial for accurately tracking small objects.
    \item We introduce a motion-aware fusion mechanism that combines motion attention maps with high-level visual features through element-wise multiplication, significantly improving tracking performance.
    \item We show that integrating motion concepts with a simple, plug-and-play fusion module into TrackNetV2 and TrackNetV3 enhances the tracking of fast-moving, small objects, resulting in our improved model, TrackNetV4.
\end{enumerate}

This paper is organized as follows: Sec.~\ref{sec:approach} introduces our motion-aware fusion framework, Sec.~\ref{sec:exp} shows our experiments and discussions, and Sec.~\ref{sec:concl} concludes the paper.

\section{Approach}
\label{sec:approach}

% In this section, we begin with an introduction to the preliminary background, followed by the presentation of our proposed motion-aware fusion, a fundamental building block that can be integrated into the existing TrackNet family.

\subsection{Preliminary}

\textbf{Notation.} Let $\idx{T}$ denote the index set ${1, 2, \cdots, T}$. We use regular fonts for scalars (\eg, $x$), lowercase boldface letters (\eg, $\vx$) for vectors, uppercase boldface letters (\eg, \boldsymbol{$\mX$}) for matrices, and calligraphic letters (\eg, $\tX$) for tensors. 
Let $\tX \in \mbr{d_1 \times d_2 \times d_3}$ denote a third-order tensor. 
Using Matlab notation, we refer to its $t$-th slice as $\tX_{:,:,t}$, which is a $d_1 \times d_2$ matrix.

\textbf{Motivation.} Prompts can extend beyond text or signals; they can also be learnable~\cite{xu2024finepose, wang2024vilt} and take various forms~\cite{NEURIPS2023_656678aa, duan2024cross, wang2024taylor, wang2024high, wang2024flow, fang2024signllm,chen2024sato}. Recent work~\cite{chen2024motion} introduces a motion prompt layer with only two learnable parameters that modulate frame differencing maps to produce motion attention maps. These maps spatially highlight regions where motion is relevant (\eg, balls) and suppress irrelevant motion (\eg, video noise and background movement), while also capturing the temporal evolution of these attention maps over time.
We explore the application of motion attention maps in tracking and predicting high-speed, small objects in professional ball games (\eg, badminton, tennis, football, golf, \etc).

\subsection{Motion Attention Maps and Motion-Aware Fusion}

Fig.~\ref{fig:main} provides an overview of our motion-aware fusion framework. For simplicity, we use TrackNetV2 for visualization, but our motion-aware fusion can be seamlessly integrated into any existing heatmap-based detection and tracking framework. Our approach offers a straightforward yet effective solution: (i) using a motion prompt layer to highlight relevant motions as attentions, and (ii) fusing the motion attention maps with high-level visual feature maps to preserve both the visual and motion information of small objects, thereby enhancing detection and tracking. Following best practices in~\cite{Sun2020TrackNetV2ES, 10.1145/3595916.3626370}, we also use a multiple-input, multiple-output design.

\begin{figure}[t]%htbp % left bottom right top
\centering%%%%
    \begin{tabular}[t]{cc}
        \subfigure[$\tD_t$ as input]
        {\label{fig:qx-pn}\includegraphics[trim=0cm 0cm 0cm 0cm, clip=true, width=0.48\linewidth]{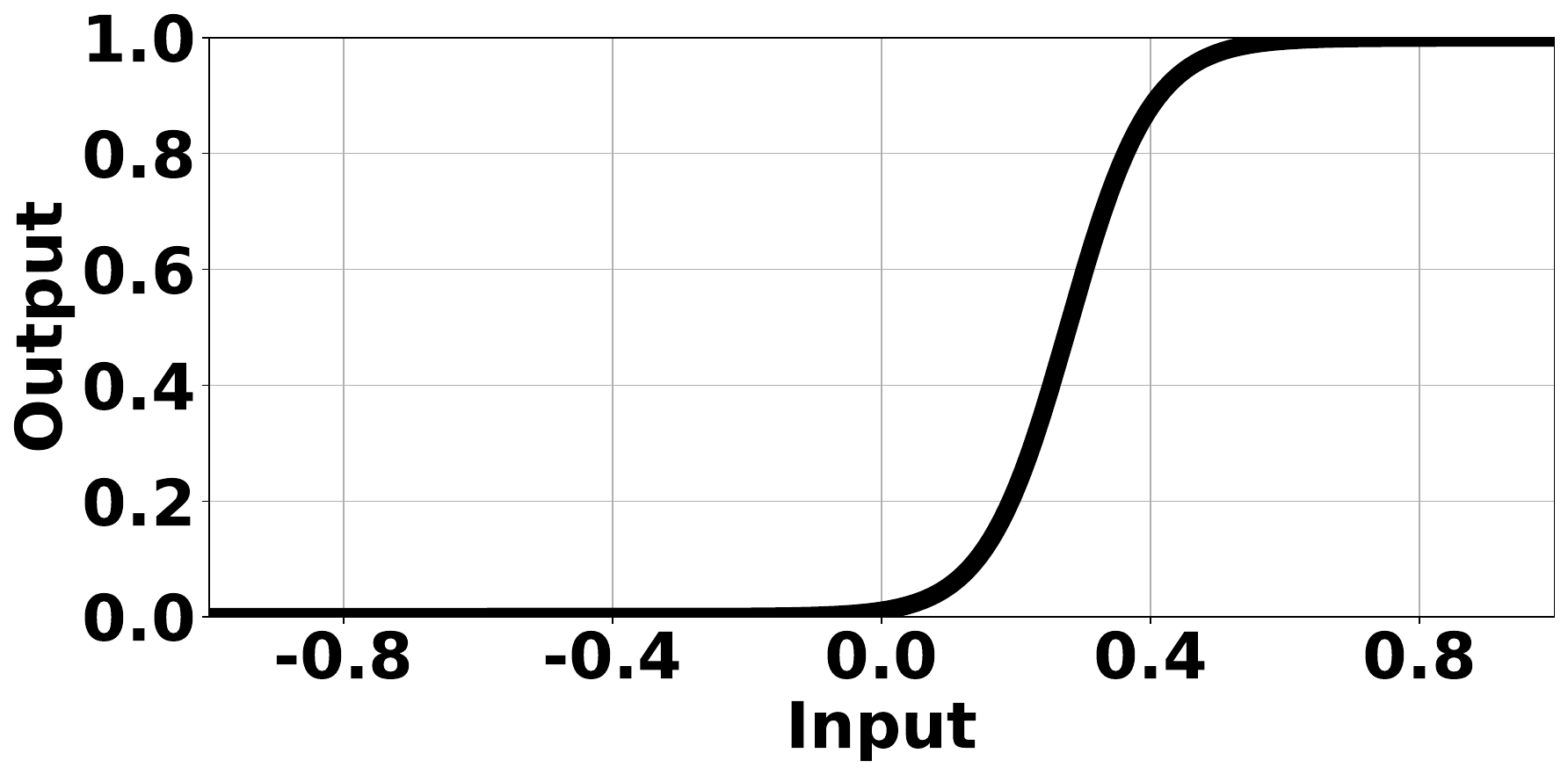}}
        \subfigure[$\tD_t^{+}$ as input (ours)]
        {\label{fig:our-pn}\includegraphics[trim=0cm 0cm 0cm 0cm, clip=true, width=0.48\linewidth]{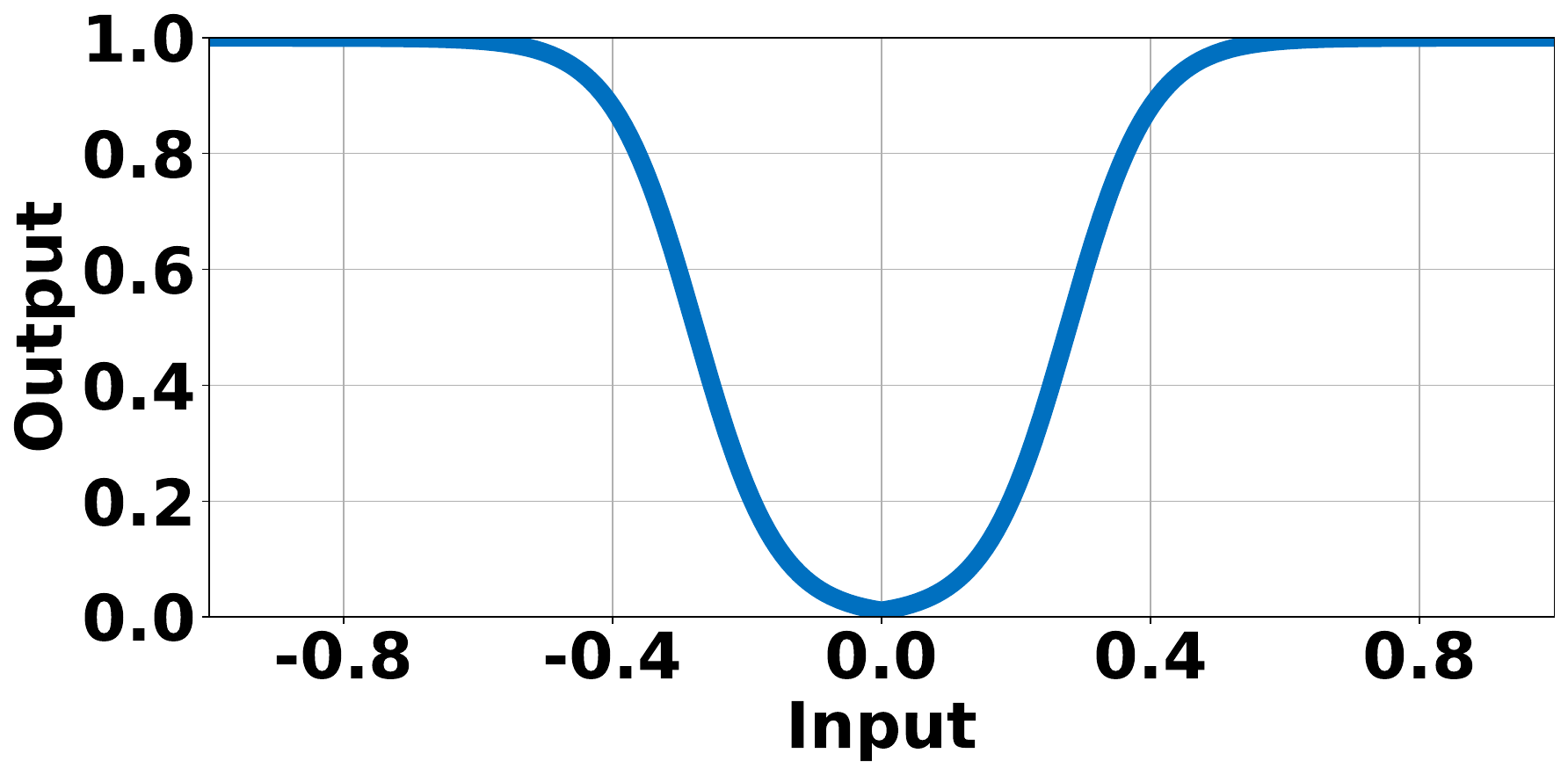}}
    \end{tabular}
\caption{Comparison between (a) original frame differencing maps $\tD_t$ and (b) absolute frame differencing maps $\tD_t^{+}$, both using the same normalization function from~\cite{chen2024motion} with a slope of 16.24 and a shift of 0.28 for visualization. Our approach captures both positive and negative intensity changes, ensuring key motions for tracking and prediction are not missed, unlike~\cite{chen2024motion}, which maps negative values to 0.} 
\label{fig:pn}
\end{figure}

\textbf{Learnable attention maps for highlighting motion.}
Given a video sequence $\tX=[\tF_1, \tF_2, \cdots, \tF_T]\in\mbr{H\times W\times 3\times T}$, where $\tF_t$ ($t \in \idx{T}$) represents the $t$-th frame, and $H$ and $W$ denote the frame height and width, respectively, we select $T'$ as the number of input frames to form short-term temporal blocks. The $t$-th temporal block is represented as $\tX_t =[\tF_t, \tF_{t+1}, \cdots, \tF_{T'+t-1}]\in\mbr{H\times W\times 3\times T'}$ for $t\in\idx{T-T'+1}$.

Each temporal block is then converted into a grayscale video sequence, $\tX'_t=[\mF'_t, \mF'_{t+1}, \cdots, \mF'_{T'+t-1}]\in\mbr{H\times W\times T'}$. After normalizing pixel values between 0 and 1, frame differencing maps are computed between consecutive frames, resulting in $\tD_t=[\mD_t, \mD_{t+1}, \cdots, \mD_{T'+t-2}]\in\mbr{H\times W\times {(T'-1)}}$, where $\mD_t=\mF'_{t+1}-\mF'_t$ ($t\in\idx{T'-1}$). Positive values in $\mD_t$ indicate an increase in pixel intensity from frames $t$ to $t+1$, while negative values indicate a decrease. 
Since $\tD_t$ includes both positive and negative pixel intensity changes, such as minor object movements, we take the absolute values to capture all relevant motions for tracking and prediction, resulting in $\tD_t^+$ ranging between 0 and 1. Fig.~\ref{fig:pn} shows the difference between using original frame differencing maps and absolute frame differencing maps. 
We apply a Power Normalization (PN) function $a$ with learnable parameters $\boldsymbol{\theta}$ as in~\cite{chen2024motion} to these differencing maps, producing a sequence of motion attention maps for the $t$-th temporal block:
\begin{equation}
    \tA_t = a_{\boldsymbol{\theta}}(\tD_t^+), 
    \label{eq:attn}
\end{equation}
where $t \in\idx{T' - 1}$ and $\tA_t\in\mbr{H\times W\times (T'-1)}$. 

\textbf{Fusing motion attention maps with visual features.} Inspired by the performance gains achieved through skip connections that address the gradual loss of tiny object features along the processing pipeline (\eg, in TrackNetV2), we introduce a specialized motion-aware fusion mechanism. This mechanism integrates high-level visual features with our motion attention maps, preserving the ball's location and trajectory.

\begin{table*}[htbp]
\caption{Performance and speed comparisons of baselines versus TrackNetV4 (baselines with our motion-aware fusion, denoted as +Motion) on the Shuttlecock and Tennis Ball tracking datasets. For Tennis Ball tracking, we apply our predefined (i) game-level and (ii) clip-level evaluation protocols. The ``Total'' column indicates the total number of frames used for evaluation. For TrackNetV2, a 3-in 3-out setup is used. The rows highlighted in blue denote our TrackNetV4.}
\vspace{-0.5cm}
\begin{center}
\resizebox{\linewidth}{!}{\begin{tabular}{l l c c c c c c c c c c c c}
\toprule
 & \multirow{2}{*}{\bf Method} & \multirow{2}{*}{\bf Total}  & \multicolumn{5}{c}{\bf Confusion matrix} & & \multicolumn{4}{c}{\bf Performance} & \multirow{2}{*}{\bf Speed} \\
\cline{4-8}
\cline{10-13}
& & & TP & TN & FP1 & FP2 & FN & & Acc. & Prec. & Rec. & F1 & \\
\midrule
\multirow{2}{*}{\parbox{2.0cm}{\bf Tennis ball (i)}} 
& TrackNetV2 & 17193 & 15863 & 396 & 142 & 17 & 775 & & 94.6 & \textbf{99.0} & 95.3 & 97.1 & 156.9\\
& \cellcolor{LightCyan}{TrackNetV2 (+Motion)} & \cellcolor{LightCyan}{17193} & \cellcolor{LightCyan}{15973} & \cellcolor{LightCyan}{389} & \cellcolor{LightCyan}{167} & \cellcolor{LightCyan}{24} & \cellcolor{LightCyan}{640} & \cellcolor{LightCyan}{} & \cellcolor{LightCyan}{\textbf{95.2}} & \cellcolor{LightCyan}{98.8} & \cellcolor{LightCyan}{\textbf{96.1}} & \cellcolor{LightCyan}{\textbf{97.5}} & \cellcolor{LightCyan}{155.7}\\ % V1_NF_RIO_10u, 0.1a, 0b, e17
\hline
\multirow{2}{*}{\parbox{2.0cm}{\bf Tennis ball (ii)}}
& TrackNetV2 & 17769 & 16195 & 393 & 163 & 25 & 993 & & 93.4 & \textbf{98.9} & 94.2 & 96.4 & 160.9\\
& \cellcolor{LightCyan}{TrackNetV2 (+Motion)}
& \cellcolor{LightCyan}{17769} & \cellcolor{LightCyan}{16374} & \cellcolor{LightCyan}{399} & \cellcolor{LightCyan}{199} & \cellcolor{LightCyan}{19} & \cellcolor{LightCyan}{778} & \cellcolor{LightCyan}{} & \cellcolor{LightCyan}{\textbf{94.4}} & \cellcolor{LightCyan}{98.7} & \cellcolor{LightCyan}{\textbf{95.5}} & \cellcolor{LightCyan}{\textbf{97.0}} & \cellcolor{LightCyan}{158.6}\\ % last conv (a=0.1,b=0,V1\_NF\_RIO\_1m, e20) 
\hline
\multirow{8}{*}{\parbox{2.0cm}{\bf Shuttlecock}}
& YOLOv7 & - & - & - & - & - & - & & 57.8 & 78.5 & 60.0 & 68.0 & - \\
& TrackNetV2 (3 in 1 out)$^\dagger$ & 13064 & 9447 & 1514 & 751 & 218 & 1134 & & 83.9 & 90.7 & 89.2 & 89.9 & \it{12.9} \\
& TrackNetV2 (3 in 3 out)$^\dagger$ & 39192 & 29129 & 4264 & 468 & 358 & 4973 & & 85.2 & 97.2 & 85.4 & 90.9 & \it{31.8} \\
& TrackNetV2 & 37794 & 26324 & 6013 & 438 & 493 & 4526 &  & 85.6 & \textbf{96.6} & 85.3 & 90.6 & 163.3\\
& \cellcolor{LightCyan}{TrackNetV2 (+Motion)}$^\ddagger$& \cellcolor{LightCyan}{37794} & \cellcolor{LightCyan}{26592} & \cellcolor{LightCyan}{5995} & \cellcolor{LightCyan}{523} & \cellcolor{LightCyan}{511} & \cellcolor{LightCyan}{4173} & \cellcolor{LightCyan}{} & \cellcolor{LightCyan}{\textbf{86.2}} & \cellcolor{LightCyan}{96.3} & \cellcolor{LightCyan}{\textbf{86.4}} & \cellcolor{LightCyan}{\textbf{91.1}} & \cellcolor{LightCyan}{139.1} \\
& \cellcolor{LightCyan}{TrackNetV2 (+Motion)} & \cellcolor{LightCyan}{37794} & \cellcolor{LightCyan}{26878} & \cellcolor{LightCyan}{5834} & \cellcolor{LightCyan}{765} & \cellcolor{LightCyan}{672} & \cellcolor{LightCyan}{3645} & \cellcolor{LightCyan}{} & \cellcolor{LightCyan}{\textbf{86.6}} & \cellcolor{LightCyan}{94.9} & \cellcolor{LightCyan}{\textbf{88.1}} & \cellcolor{LightCyan}{\textbf{91.4}} & \cellcolor{LightCyan}{161.1} \\
& TrackNetV3 & 10836 & 8980 & 1395 & 22 & 8 & 431 & & 95.7 & \textbf{99.7} & 95.4 & 97.5 & 15.1$^*$\\
% & TrackNetV3 & 37794 & 26709 & 5990 & 439 & 490 & 4166 & & 86.3 & 96.7 & 86.5 & 91.5 & \\
& \cellcolor{LightCyan}{TrackNetV3 (+Motion)}& \cellcolor{LightCyan}{10836} & \cellcolor{LightCyan}{9050} & \cellcolor{LightCyan}{1400} & \cellcolor{LightCyan}{30} & \cellcolor{LightCyan}{10} & \cellcolor{LightCyan}{346} & \cellcolor{LightCyan}{} & \cellcolor{LightCyan}{\textbf{96.4}} & \cellcolor{LightCyan}{99.5} & \cellcolor{LightCyan}{\textbf{96.3}} & \cellcolor{LightCyan}{\textbf{97.9}} & \cellcolor{LightCyan}{15.1}$^*$ \\
\bottomrule
\end{tabular}}
\label{tab:globaltable}
\end{center}
$^\dagger$ indicates results from the original TrackNetV2 paper~\cite{Sun2020TrackNetV2ES}. $^\ddagger$ indicates the results obtained by fine-tuning the pretrained baseline TrackNetV2 for 3 epochs. 

$^*$ indicates the processing speed of the entire script, including data loading, file writing, \etc. This may not be directly comparable to the other speeds.
\end{table*}

Specifically, we first extract high-level visual feature maps using the tracking network up to the last convolutional block (just before the Sigmoid layer that outputs the heatmaps):
\begin{equation}
    \tV_t = \text{TrackNet}_\text{visual}(\tX_t),
\end{equation}
where $\text{TrackNet}_\text{visual}(\cdot)$ refers to the TrackNet for extracting the visual features $\tV_t=[\mV_t, \mV_{t+1}, \cdots, \mV_{T'+t-1}] \in \mbr{H\times W\times T'}$. We then aggregate these visual feature representations with the motion attention maps generated via Eq.~\eqref{eq:attn} from the motion prompt layer:
\begin{equation}
    \tH_t = \sigma(\tA_t \circledcirc \tV_t),
    \label{eq:new-heatmap}
\end{equation}
where $\sigma(\cdot)$ is the Sigmoid function, and $\tH_t \in\mbr{H\times W\times T'}$ denotes the motion-attention-enhanced heatmaps. The symbol $\circledcirc$ represents our fusion operation, which involves element-wise multiplication followed by concatenation:
\begin{equation}
    \tA_t \circledcirc \tV_t \! = \!\left[\mV_t, \mA_t \odot \mV_{t\!+\!1}, \cdots, \mA_{T'\!+\!t\!-\!2} \odot \mV_{T'\!+\!t\!-\!1} \right],
    \label{eq:fusion}
\end{equation}
% \begin{align}
%     & \quad \tA_t \circledcirc \tV_t \nonumber \\
%     % & = \!\left[\mV_t, \frac{1}{2}\!\!\left(\!\frac{\!\mV_t \!\!+ \!\!\mA_t}{2} \!+\! \mV_{t+1}\!\right), \cdots, \frac{1}{2}\!\!\left(\!\frac{\!\mV_{T'\!+\!t\!-\!2} \!\!+ \!\!\mA_{T'\!+\!t\!-\!2}}{2} \!+\! \mV_{T'\!+\!t\!-\!1}\!\right) \right],
%     & \! = \!\left[\mV_t, \mA_t \odot \mV_{t\!+\!1}, \cdots, \mA_{T'\!+\!t\!-\!2} \odot \mV_{T'\!+\!t\!-\!1} \right],
% \end{align}
where $\mA_\tau \!\odot\! \mV_{\tau\!+\!1}$ ($\tau \in \idx{T'+t-2}$) denotes the motion-enhanced visual representations, and $\tA_t \circledcirc \tV_t \in \mbr{H\times W\times T'}$. As shown in Eq.~\eqref{eq:new-heatmap}, these representations are then passed through a Sigmoid function to produce the final heatmaps, which highlight the ball's location and trajectory over time. 

% To further enhance performance, we incorporate a temporal attention variation regularization term, as in~\cite{chen2024motion}, which aids in the effective learning of smooth motion attentions and ensures the tracking system maintains a refined focus on relevant motions. In sports videos, where balls are typically small and move at high speeds, we use smaller regularization parameters (\eg, $1e-5$, $1e-3$) to prevent important information from being lost as noise.

We name the tracking framework incorporating our motion-aware fusion mechanism TrackNetV4. Below, we present our experimental results, comparisons and discussions.

% \begin{figure}[t]
% \centering
%     \includegraphics[trim=0cm 0cm 0cm 0cm, clip=true, width=\linewidth]{imgs/feature_maps_predictions/main_visualization.pdf}
% \caption{Comparison of feature maps and heatmaps with and without motion-aware fusion using TrackNetV2 as the backbone. The first row shows the original frames. Motion-aware fusion enhances visual representations (\eg, second row vs. third row), resulting in clearer and more accurate ball location predictions. Additionally, when combined with high-level features, our motion attention maps further improve ball localization, as demonstrated in the heatmaps (\eg, fourth row vs. fifth row), reducing missed detections compared to TrackNetV2. This highlights that incorporating motion awareness significantly enhances the model's ability to track high-speed, small objects.}
% \label{fig:heat-vis}
% \end{figure}

\begin{figure}[t]
\subfigure[Improved feature maps.]
    {\label{fig:tennisd}\includegraphics[trim=0cm 0cm 0cm 0cm, clip=true, width=0.48\linewidth]{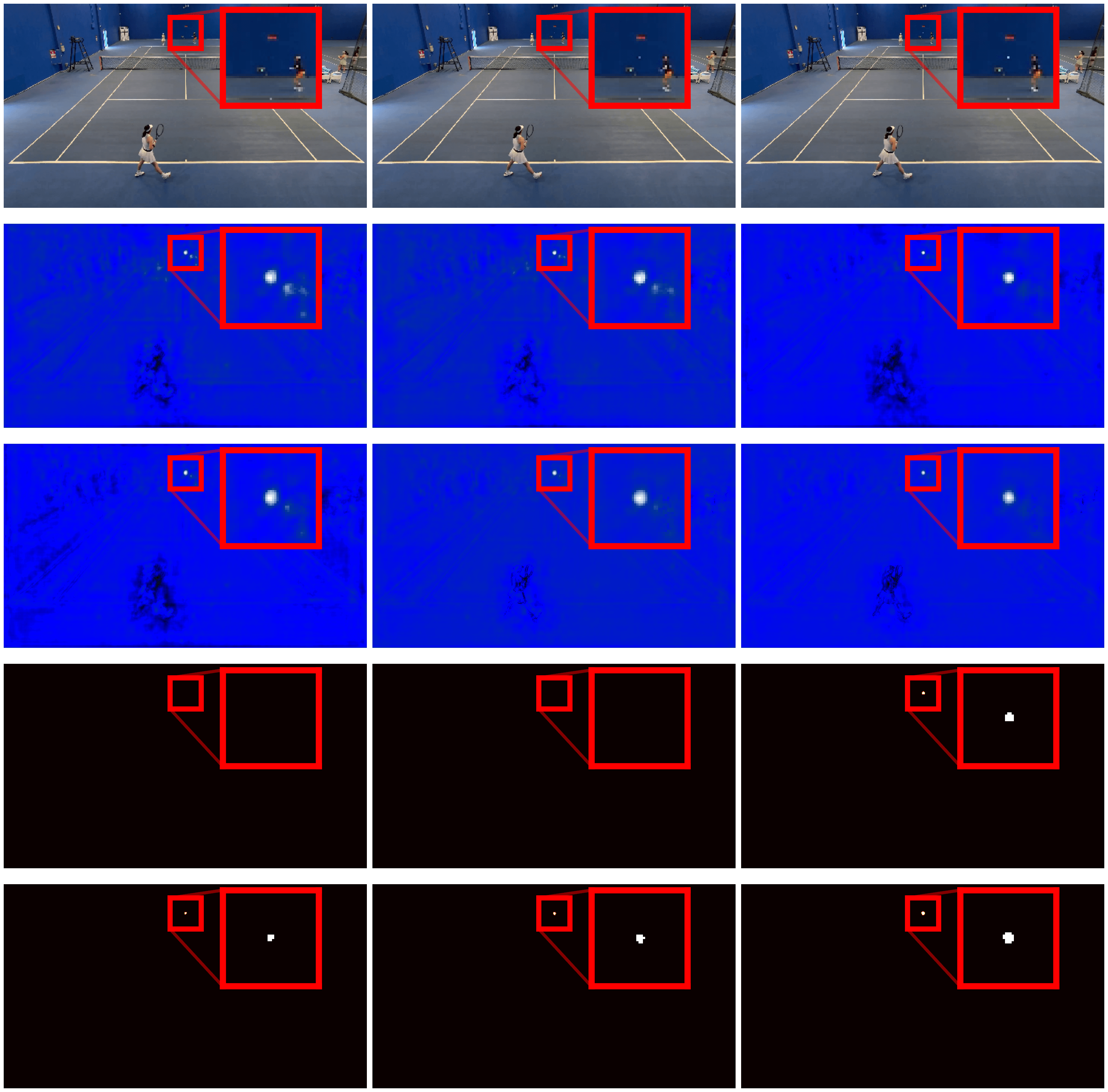}}\hfill
    \subfigure[Clearer ball predictions.]
    {\label{fig:tennisb}\includegraphics[trim=0cm 0cm 0cm 0cm, clip=true, width=0.48\linewidth]{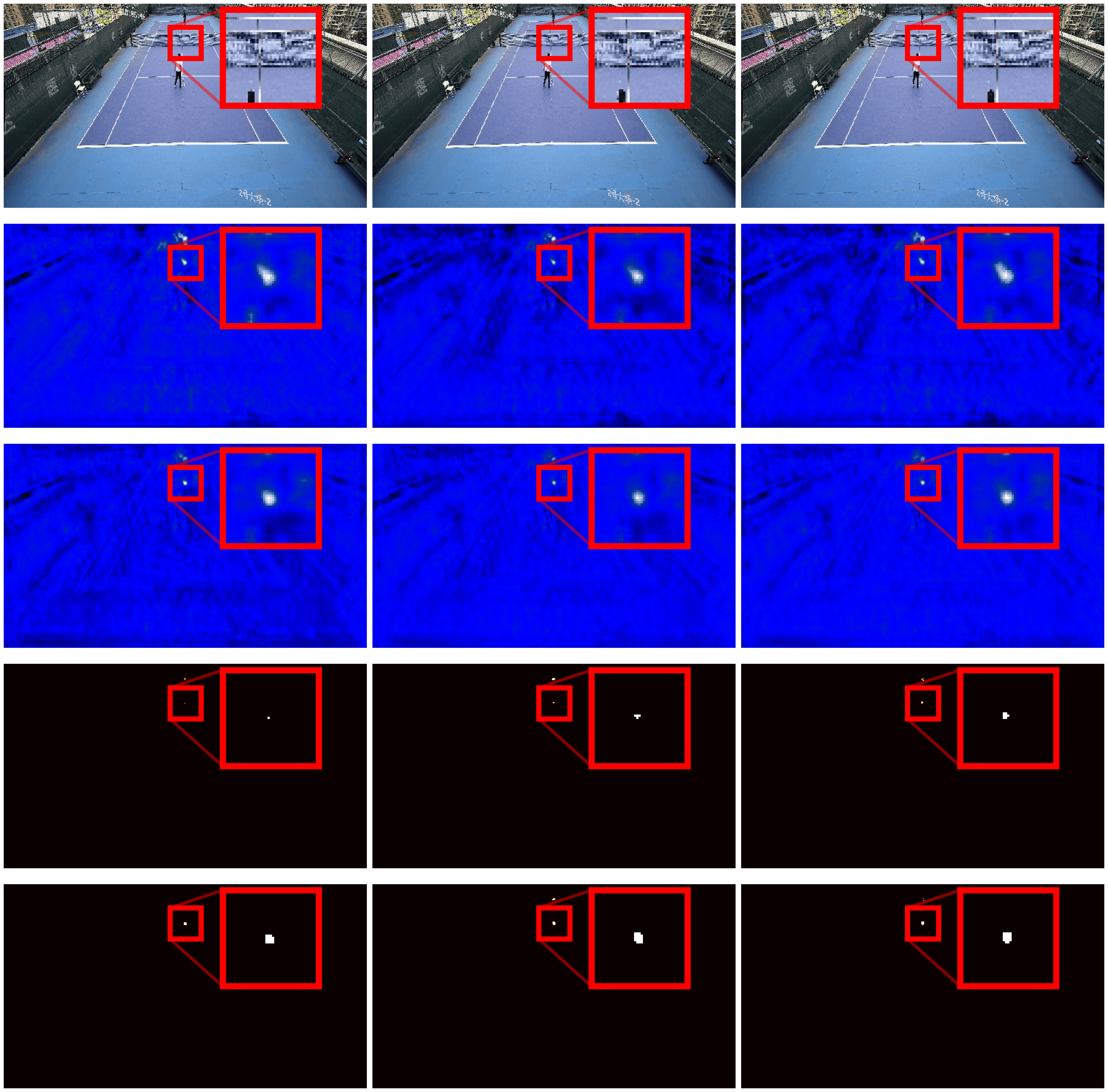}}

    \subfigure[Refined ball localization.]
    {\label{fig:tennisc}\includegraphics[trim=0cm 0cm 0cm 0cm, clip=true, width=0.48\linewidth]{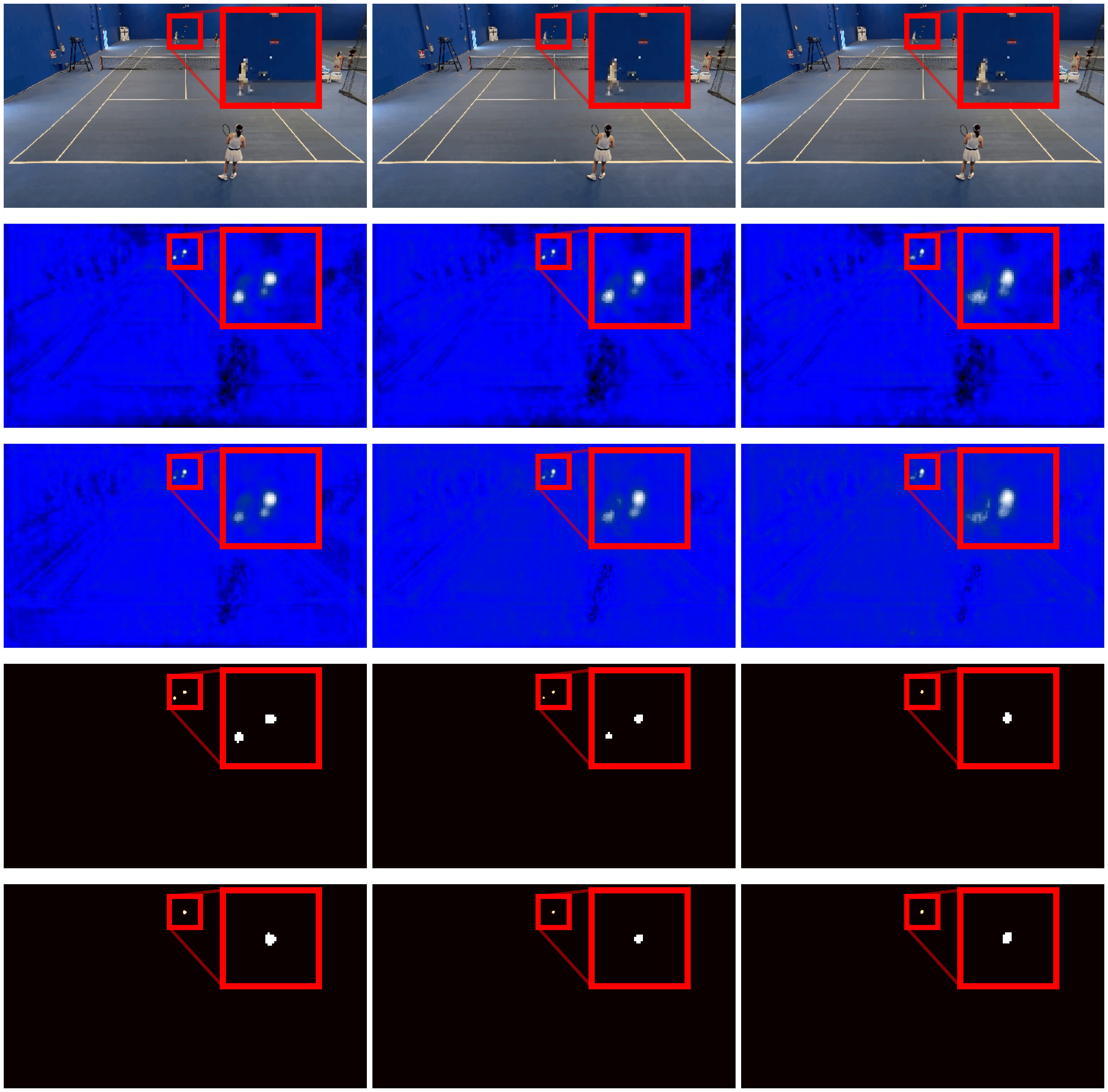}}\hfill
    \subfigure[Reduced missed detections.]
    {\label{fig:tennisa}\includegraphics[trim=0cm 0cm 0cm 0cm, clip=true, width=0.48\linewidth]{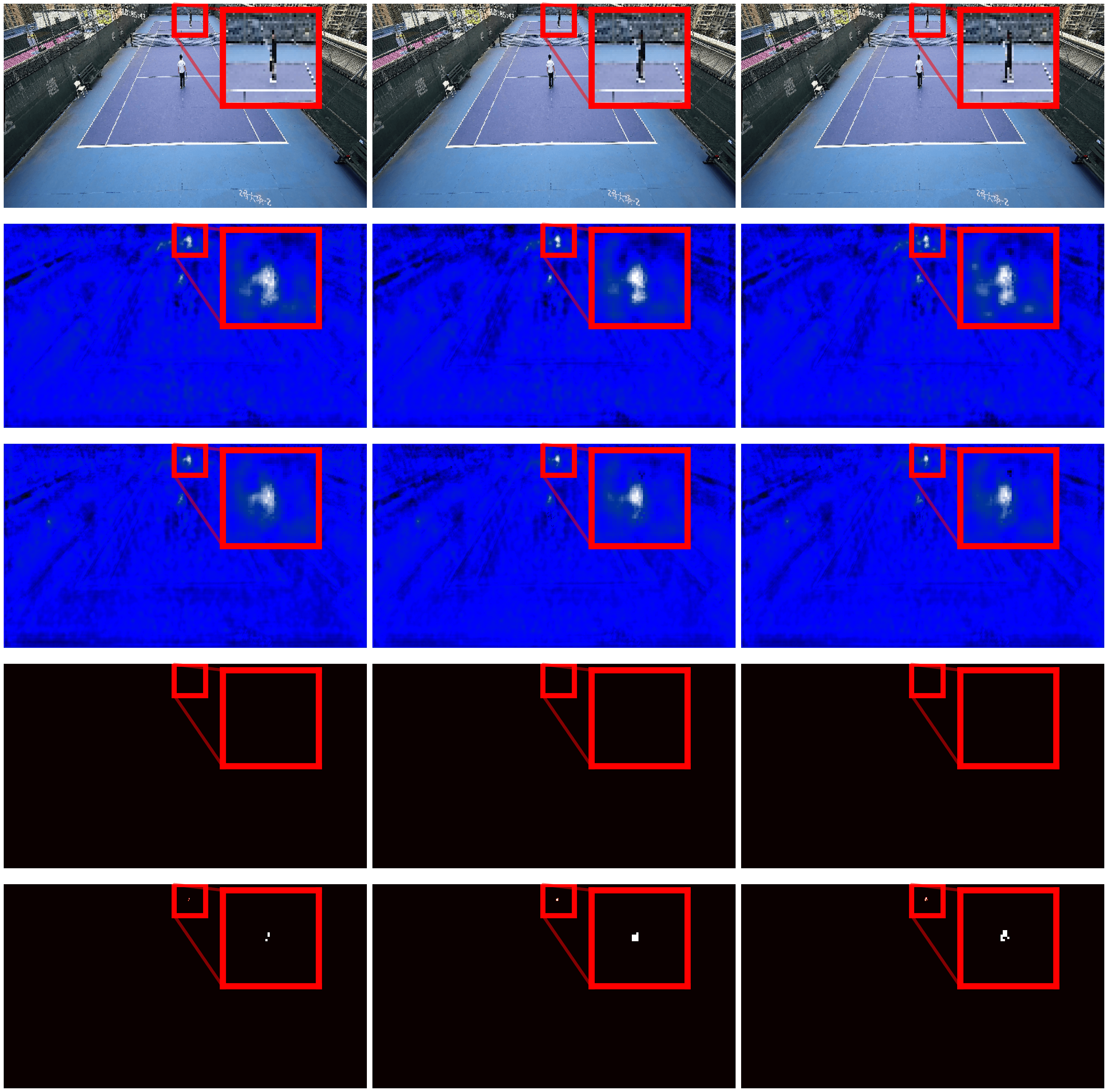}}
    
\caption{Comparison of feature maps and heatmaps with and without motion-aware fusion.
Four visualization groups are shown, with the first row in each displaying the original frames. Motion-aware fusion improves visual representations (\eg, 2nd vs. 3rd row in (a)), resulting in clearer, more accurate ball predictions ((a) and (c)). Combined with high-level features, motion attention further refines ball localization (\eg, 4th vs. 5th row in (c)), reducing missed detections compared to the baseline ((b) and (d)). This demonstrates how motion awareness enhances tracking of fast-moving, small objects.}
\label{fig:heat-vis}
\end{figure}

\section{Experiment}
\label{sec:exp}

\textbf{Dataset, protocol, and setup.}
The tennis ball tracking dataset introduced in~\cite{huang2019tracknet} lacks a standardized training and test split, as it is initially divided randomly. This random division has led to inconsistencies across different works, making it difficult to fairly compare the results with other models. To address this issue, we develop two evaluation protocols while maintaining the 70/30 frame split: 
\renewcommand{\labelenumi}{\roman{enumi}.}
\begin{enumerate}[leftmargin=0.4cm]
    \item \textit{Game-level}: The training set includes `game5', `game10', `game6', `game2', `game7', `game3', and `game8', , while the test set comprises `game1', `game9', and `game4'. This division results in 70.81\% of the total frames being used for training and 29.19\% for testing.
    \item \textit{Clip-level}: The dataset, originally organized into several clips per game, is split based on cumulative frames to ensure that 70\% of the total frames are allocated for training. Each clip is assigned entirely to either the training or test set, maintaining disjoint training and test sets.
\end{enumerate}

Additionally, we train and test our framework on the shuttlecock dataset introduced in~\cite{Sun2020TrackNetV2ES}, following the standard evaluation protocol.
For all evaluations, we use the same loss functions and parameter setups~\cite{Sun2020TrackNetV2ES, 10.1145/3595916.3626370} as the original authors, including training epochs and learning rates.

\textbf{Evaluation metrics.} We evaluate our framework's performance by measuring the distance between the predicted and actual positions of the ball. Following the criteria set in~\cite{Sun2020TrackNetV2ES,10.1145/3595916.3626370}, a detection is considered accurate if this distance is within 4 pixels; otherwise, it is deemed inaccurate. Additionally, any inconsistency in detecting the ball's presence or absence is also classified as inaccurate. Beyond accuracy, we report precision, recall, and F1 score to provide a comprehensive evaluation. To assess the efficiency of our framework, we also measure the frames per second (FPS).

\textbf{Qualitative results.}
Fig.~\ref{fig:heat-vis} presents a visual comparison of (i) standard visual feature maps from the baseline (TrackNetV2) versus motion-enhanced feature maps obtained by applying our method (using element-wise multiplication as described in Eq.~\eqref{eq:fusion}), and (ii) heatmaps generated by the baseline (TrackNetV2) compared to those produced by our TrackNetV4 (TrackNetV2 + motion-aware fusion).

We observe that the motion-aware fusion significantly enhances the tracking and prediction of ball locations, as demonstrated by the clearer visualizations of both the feature maps (before applying the Sigmoid function) and the generated heatmaps. Notably, our feature maps exhibit greater clarity, and the resulting heatmaps demonstrate increased robustness. This highlights the effectiveness of our method, which remains lightweight while successfully tracking high-speed, small objects in sports scenarios.

\textbf{Quantitative results.} Table~\ref{tab:globaltable} summarizes our results on both the tennis ball and shuttlecock datasets. We denote the use of our motion-aware fusion as `(+Motion)' in the table. 
As shown, TrackNetV4 (applying motion attention maps to TrackNetV2) consistently improves performance, particularly in accuracy and F1-score metrics, while also reducing the number of false negatives in both tennis ball evaluation protocols.

We also observe that, in general, protocol (i) game-level evaluations outperform protocol (ii) clip-level evaluations across all four metrics by more than 0.5\%. This suggests that models trained on game-level videos perform better, likely due to the scene-dependent nature of sports activities.

For the shuttlecock dataset, performance is generally lower compared to the tennis ball dataset, likely because the tennis ball is larger, more uniformly textured, and rounder in shape. However, our motion-aware fusion mechanism still achieves improvements of over 0.8\%, highlighting the importance of incorporating motion for enhanced tracking and prediction.

\section{Conclusion}
\label{sec:concl}

In this paper, we present TrackNetV4, an advanced tracking framework that integrates motion-aware fusion to enhance the tracking and prediction of fast-moving, small objects in sports videos. TrackNetV4 builds on existing tracking technologies, incorporating a novel fusion mechanism that significantly improves performance in challenging conditions, such as occlusions and limited visibility. Our extensive experimental results highlight substantial performance improvements over the previous TrackNetV3, showcasing TrackNetV4's superiority in high-speed object tracking. Notably, TrackNetV4’s lightweight and modular design allows for easy integration into various applications. % Future research will focus on applying TrackNetV4 to additional sports domains and tracking scenarios, as well as evaluating its performance with different backbone networks to further establish its generalizability and robustness.

\section*{Acknowledgment}

Arjun Raj conducted this research under the supervision of Lei Wang for his COMP3770 Computing Research Project (R\&D) at ANU. He is a recipient of research sponsorship from Active Intelligence Australia Pty Ltd in Perth, Western Australia, including The Active Intelligence Research Challenge Award. This work was also supported by the NCI National AI Flagship Merit Allocation Scheme, and the National Computational Merit Allocation Scheme 2024 (NCMAS 2024), with computational resources provided by NCI Australia, an NCRIS-enabled capability supported by the Australian Government.

\appendix

\begin{figure}[htbp]%htbp % left bottom right top
\centering%%%%
    \includegraphics[trim=5cm 0cm 5cm 0cm, clip=true, width=0.95\linewidth]{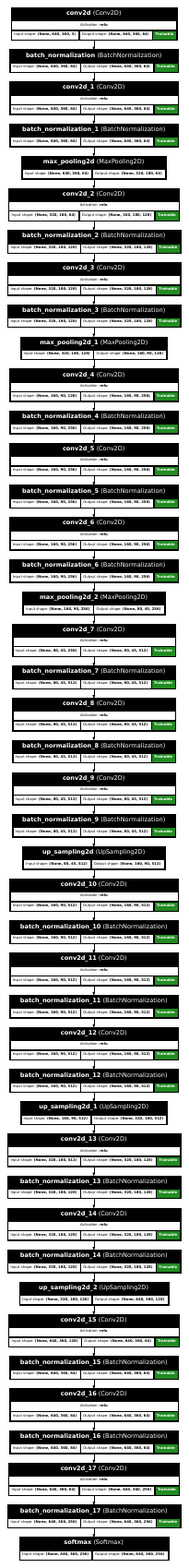}
\caption{TrackNetV1.} 
\label{fig:tracknetv1}
\end{figure}

\begin{figure}[htbp]%htbp % left bottom right top
\centering%%%%
    \includegraphics[trim=5cm 0cm 5cm 0cm, clip=true, width=0.95\linewidth]{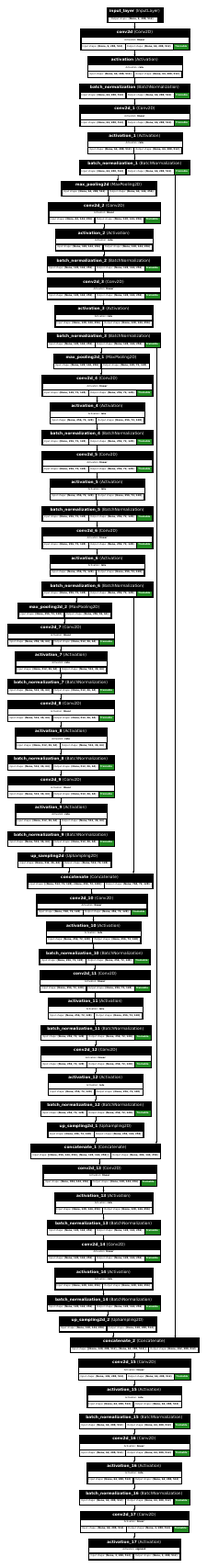}
\caption{TrackNetV2.} 
\label{fig:tracknetv2}
\end{figure}

\begin{figure}[htbp]%htbp % left bottom right top
\centering%%%%
    \includegraphics[trim=5cm 0cm 5cm 0cm, clip=true, width=0.95\linewidth]{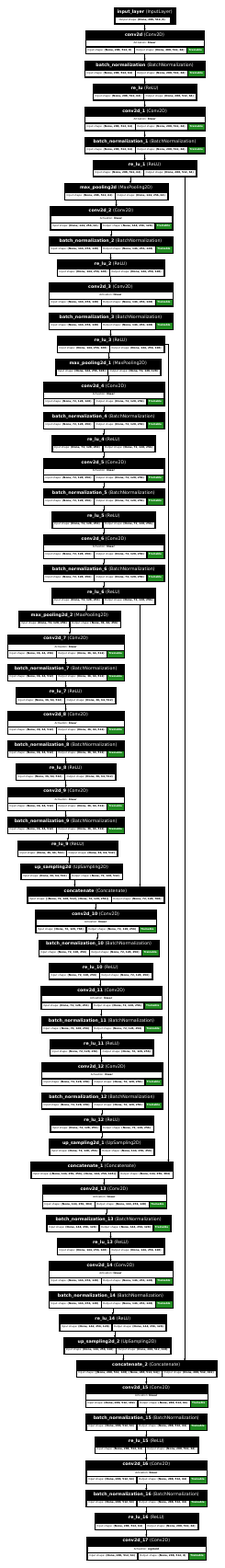}
\caption{TrackNetV3.} 
\label{fig:tracknetv3}
\end{figure}

\begin{figure*}[htbp]%htbp % left bottom right top
\centering%%%%
    \includegraphics[trim=0cm 0cm 0cm 0cm, clip=true, width=0.44\linewidth]{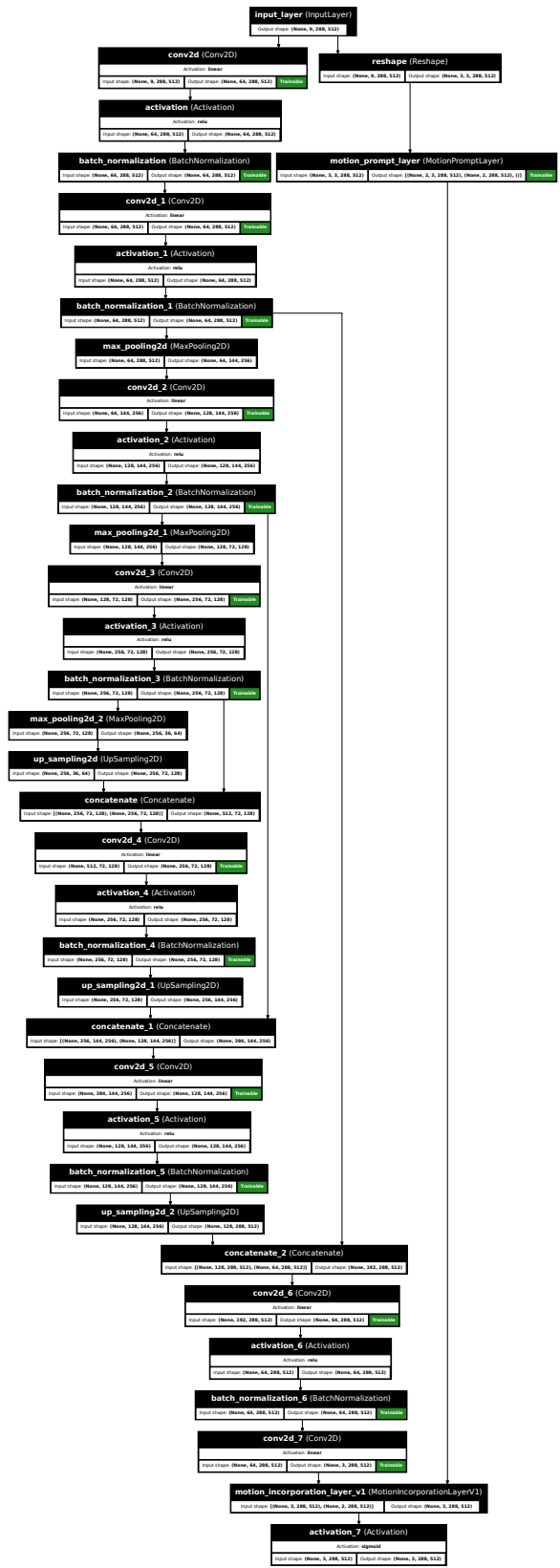}
\caption{Our TrackNetV4 model integrates modern motion concepts into the traditional 2D CNN-based TrackNet family (with TrackNetV2 chosen for visualization purposes). It enables precise, high-speed tracking of small objects in sports activities. We aim to inspire renewed interest in revisiting these older, yet still powerful models by enhancing them with contemporary motion concepts~\cite{chen2024motion}.} 
\label{fig:tracknetv4}
\end{figure*}

\begin{figure*}[htbp]
\centering
\subfigure[Singles match.]
    {\label{fig:singles-game}\includegraphics[trim=0cm 0cm 0cm 0cm, clip=true, width=0.32\linewidth]{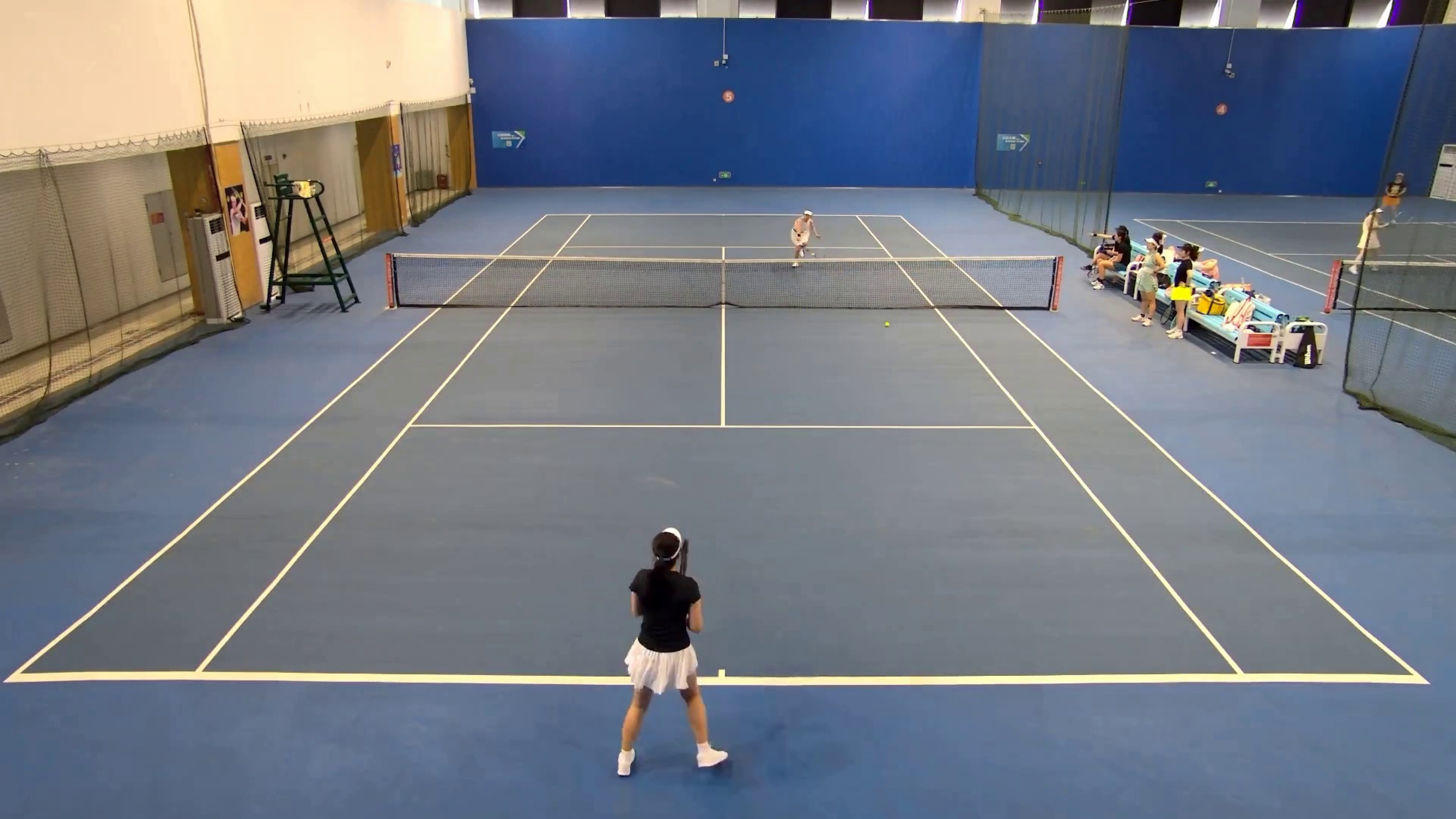}}
    \subfigure[Doubles match.]
    {\label{fig:doubles-game-1}\includegraphics[trim=0cm 0cm 0cm 0cm, clip=true, width=0.32\linewidth]{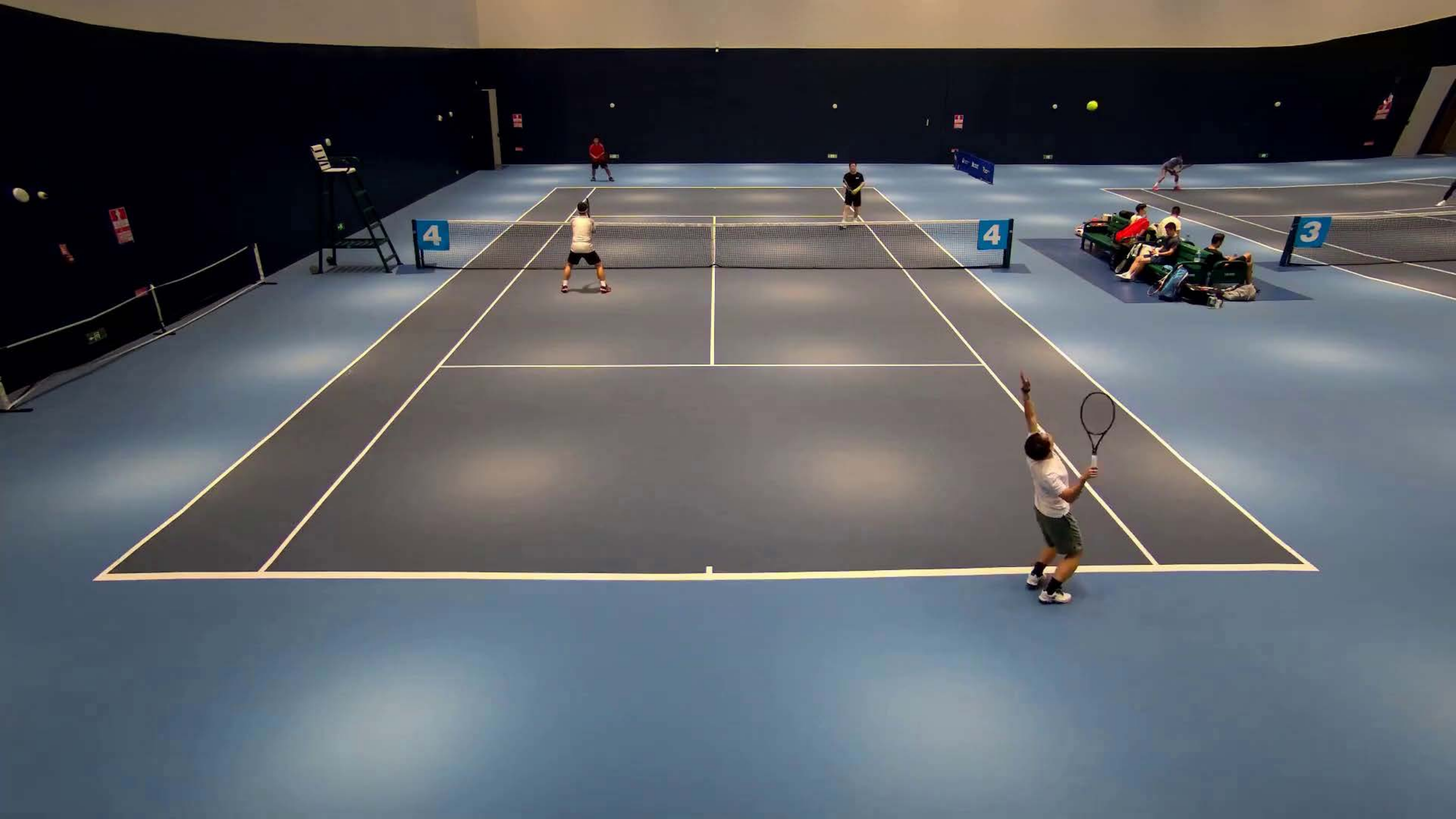}}
    \subfigure[Visible balls in play on two courts.]
    {\label{fig:multi-ball-2}\includegraphics[trim=0cm 0cm 0cm 0cm, clip=true, width=0.32\linewidth]{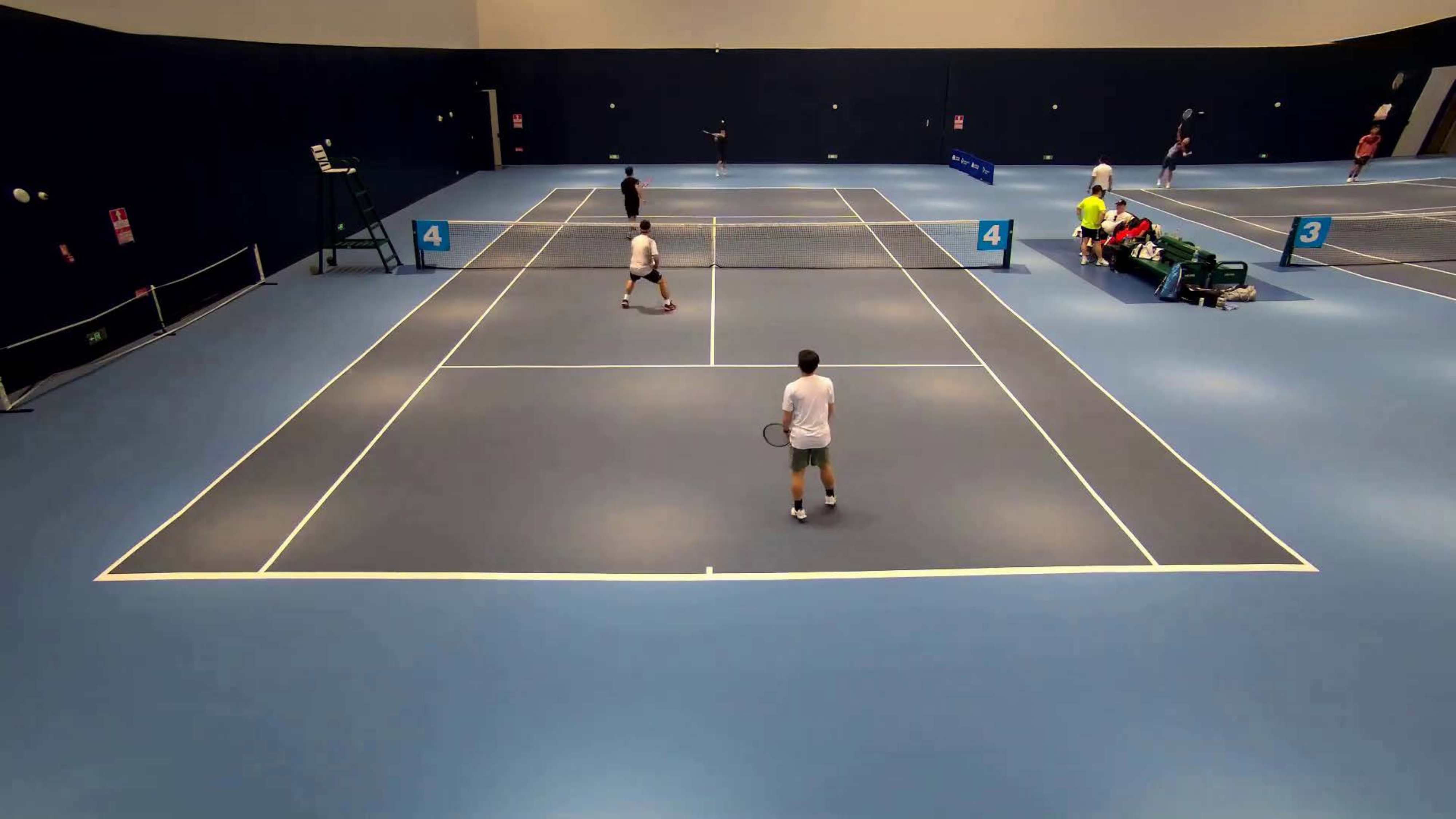}}
    \subfigure[Multiple balls in play.]
    {\label{fig:multi-ball-1}\includegraphics[trim=15cm 0cm 15cm 0cm, clip=true, width=0.24\linewidth]{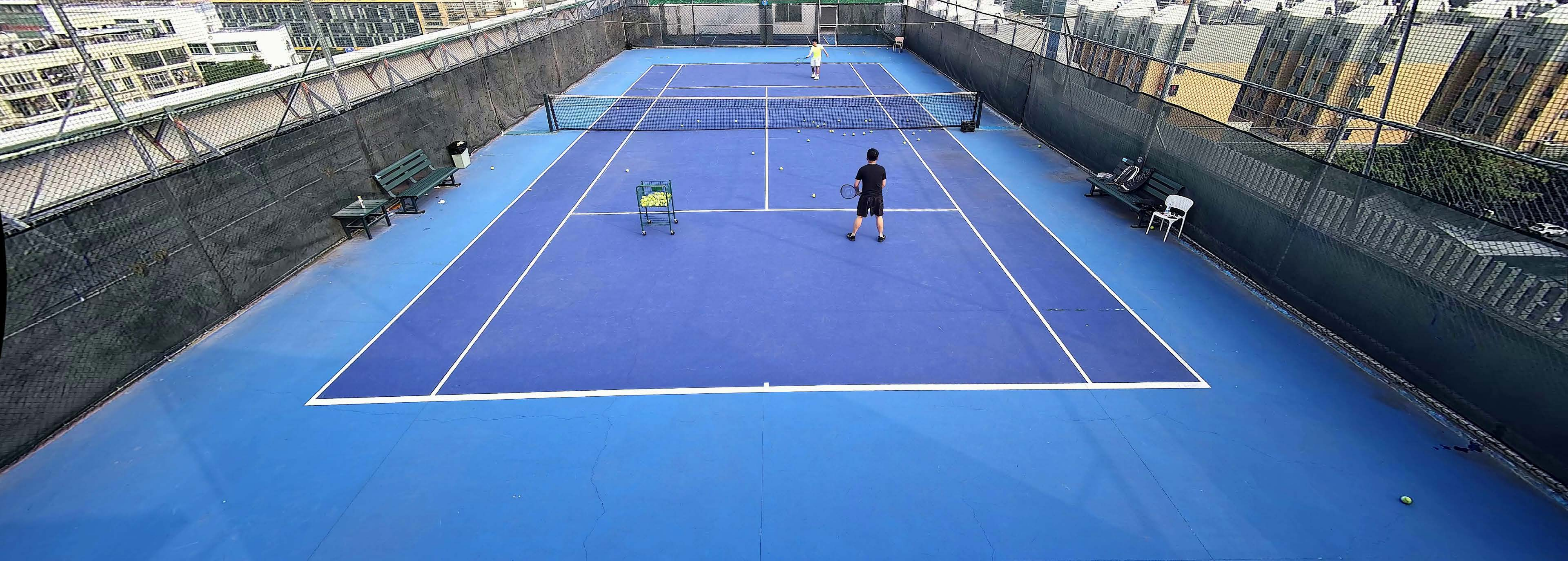}}
    \subfigure[Nighttime match.]
    {\label{fig:night-time}\includegraphics[trim=15cm 0cm 15cm 0cm, clip=true, width=0.24\linewidth]{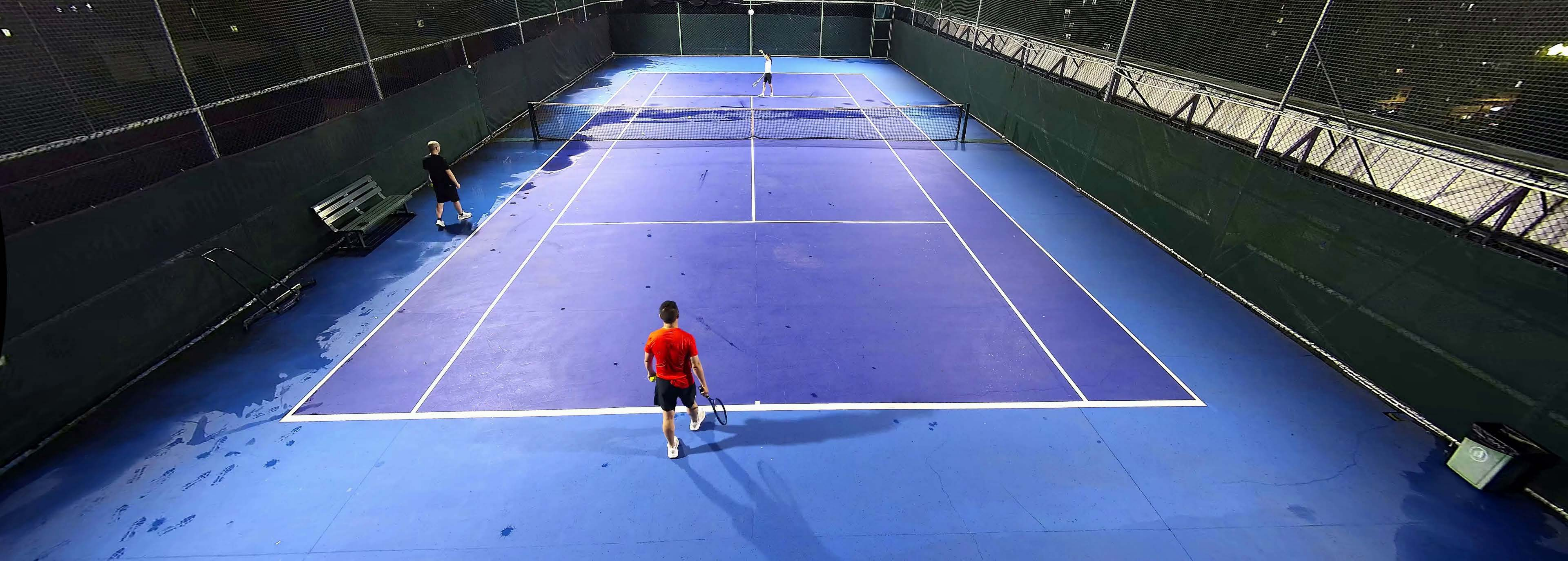}}
    \subfigure[Ball camouflages with the court.] % Ball camouflaged by the court's color.
    {\label{fig:ball-camouflaged}\includegraphics[trim=15cm 0cm 15cm 0cm, clip=true, width=0.24\linewidth]{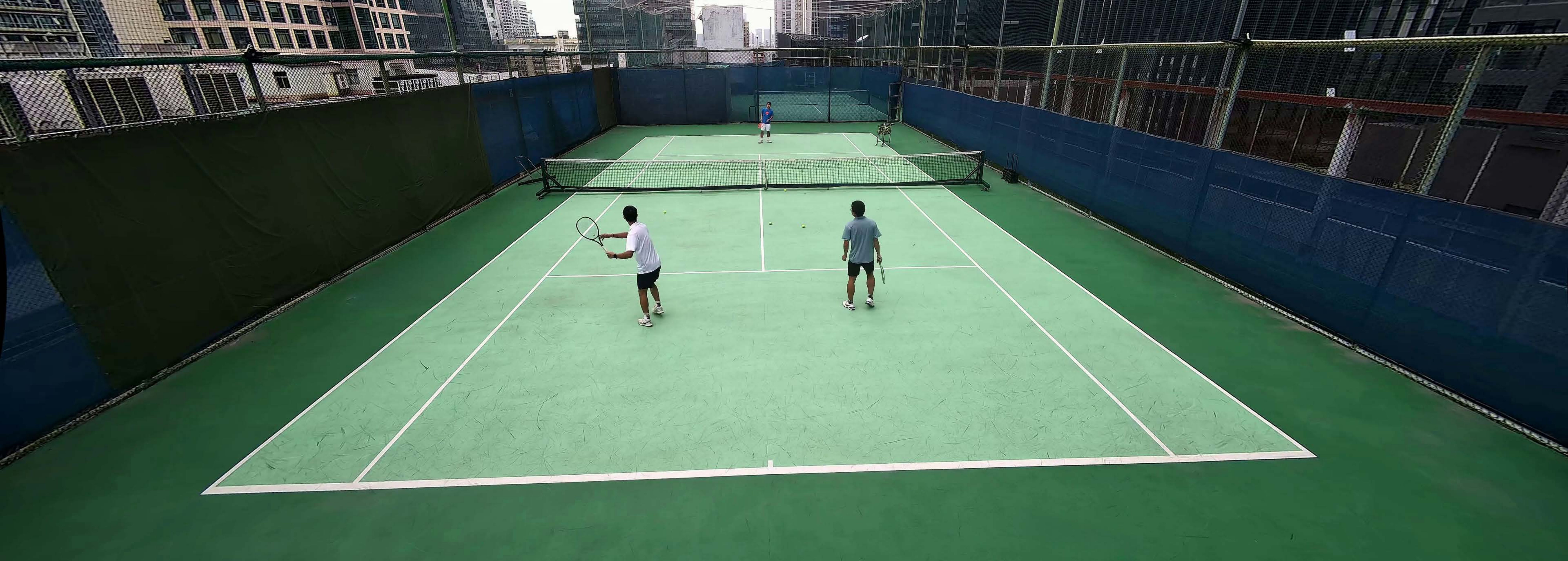}}
    \subfigure[Balls blend with the court.]
    {\label{fig:multi-ball-3}\includegraphics[trim=15cm 0cm 15cm 0cm, clip=true, width=0.24\linewidth]{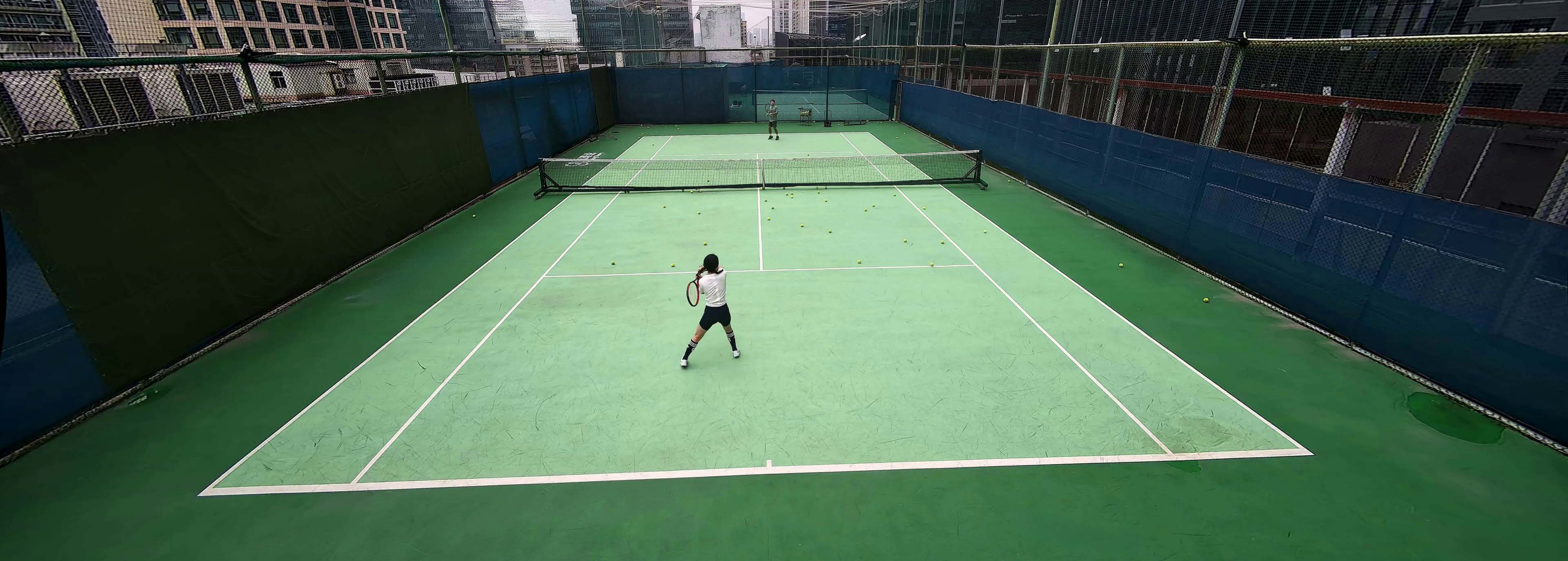}}
\caption{Our multi-ball tracking dataset includes (a) singles and (b) doubles matches, with challenges such as (c) visible balls in play on two courts within a single video, (d) multiple balls in play, and scenarios that are challenging for ball tracking, including (e) nighttime matches, (f) balls camouflaged by the court's color, and (g) balls blending into the court's color. The dataset also features a range of resolutions.}
\label{fig:internal-dataset}
\end{figure*}

\begin{figure}[htbp]
    \centering
    \includegraphics[width=\linewidth]{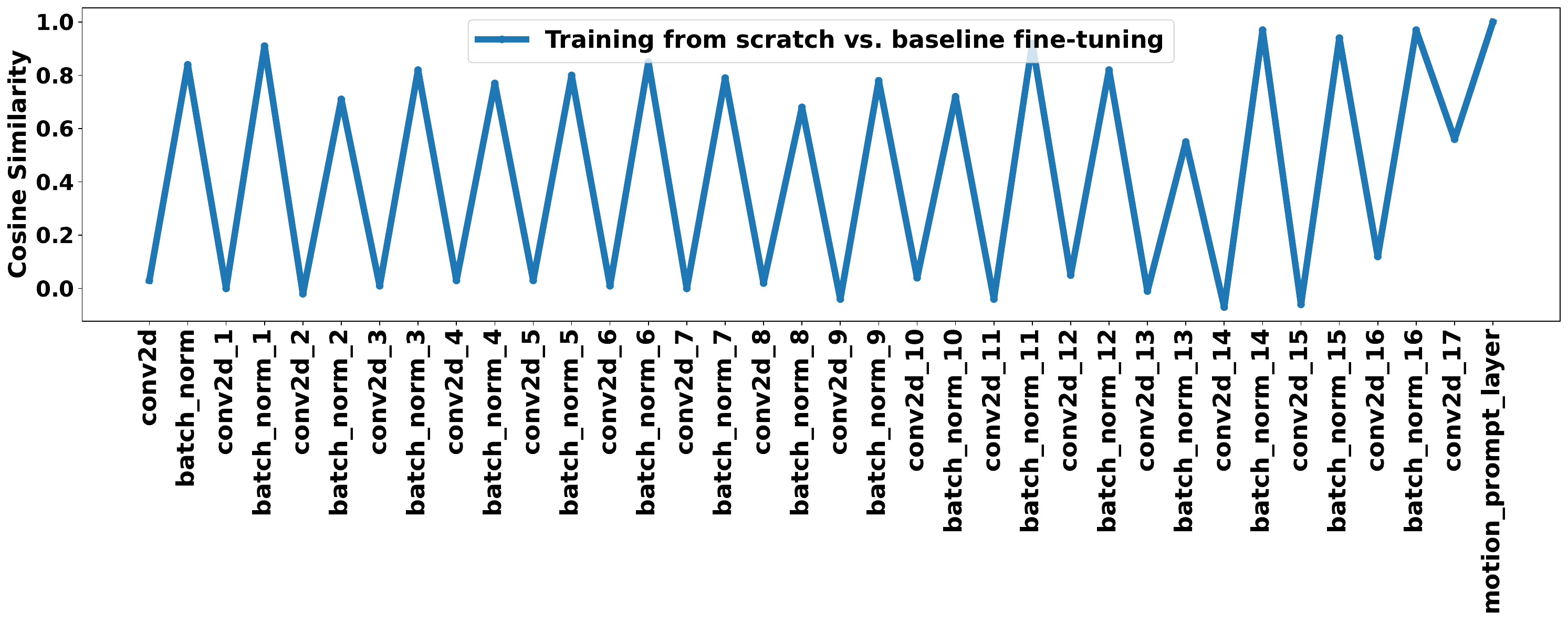}
    \caption{Per-layer weight similarity measured using cosine similarity between training from scratch and baseline fine-tuning on TrackNetV4. Note that non-trainable layers are not included. We observe significant differences in the per-layer weights, particularly in the 2D convolutional layers, despite both models achieving very similar performance.}
    \label{fig:similarity_plot}
\end{figure}

\begin{figure*}[htbp]
\centering
    \subfigure[Attention maps from the tennis ball tracking dataset~\cite{huang2019tracknet}.]
    {\label{fig:attn-map-1}\includegraphics[trim=0cm 0cm 0cm 0cm, clip=true, width=\linewidth]{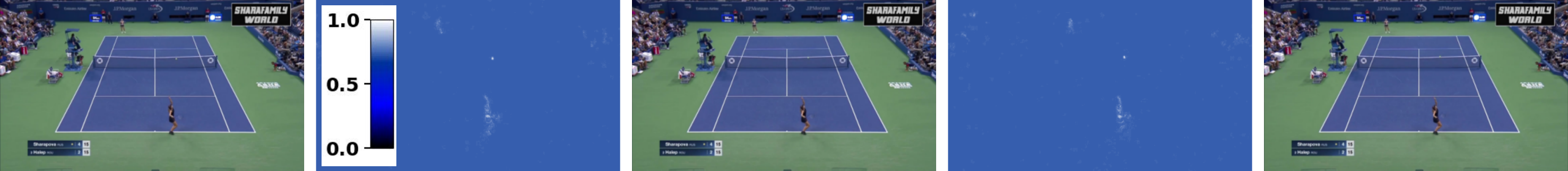}}
    \subfigure[Attention maps from the tennis ball tracking dataset~\cite{huang2019tracknet}.]
    {\label{fig:attn-map-2}\includegraphics[trim=0cm 0cm 0cm 0cm, clip=true, width=\linewidth]{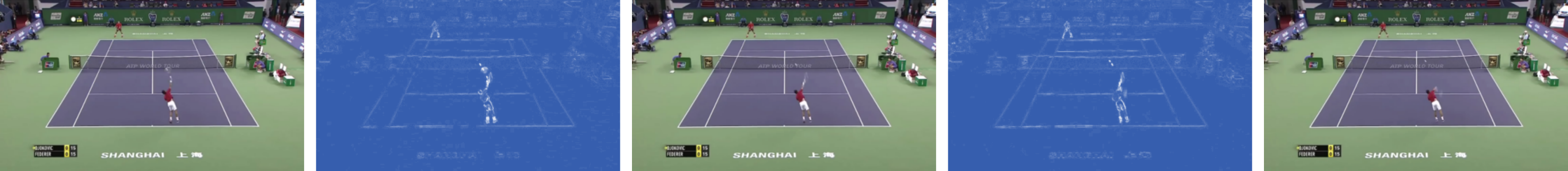}}
    \subfigure[Attention maps from the shuttlecock tracking dataset~\cite{Sun2020TrackNetV2ES}.]
    {\label{fig:attn-map-3}\includegraphics[trim=0cm 0cm 0cm 0cm, clip=true, width=\linewidth]{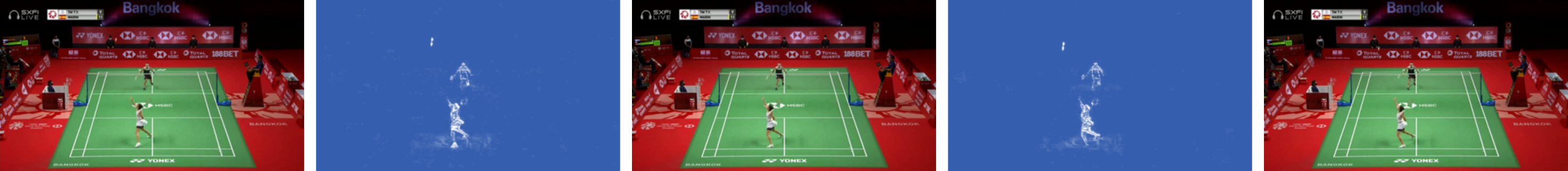}}
    \subfigure[Attention maps from the shuttlecock tracking dataset~\cite{Sun2020TrackNetV2ES}.]
    {\label{fig:attn-map-4}\includegraphics[trim=0cm 0cm 0cm 0cm, clip=true, width=\linewidth]{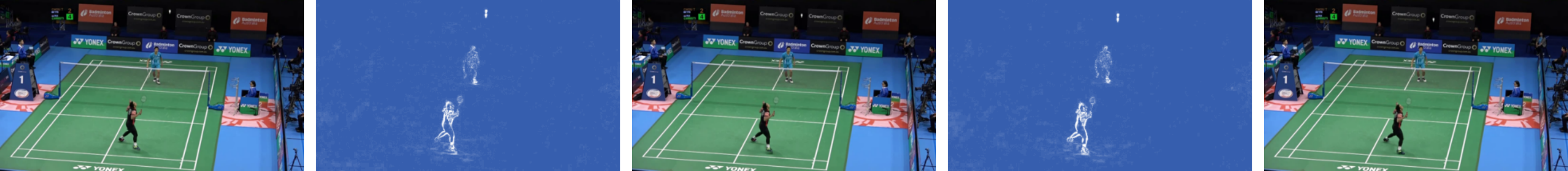}}
    \subfigure[Attention maps from the multi-ball tracking dataset.]
    {\label{fig:attn-map-5}\includegraphics[trim=0cm 0cm 0cm 0cm, clip=true, width=\linewidth]{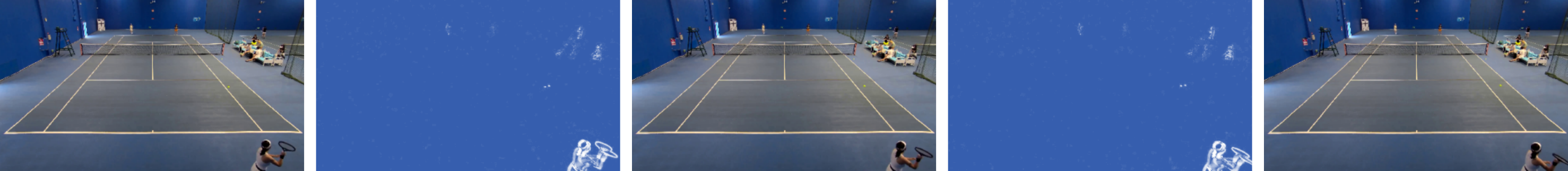}}
    \subfigure[Attention maps from the multi-ball tracking dataset.]
    {\label{fig:attn-map-6}\includegraphics[trim=0cm 0cm 0cm 0cm, clip=true, width=\linewidth]{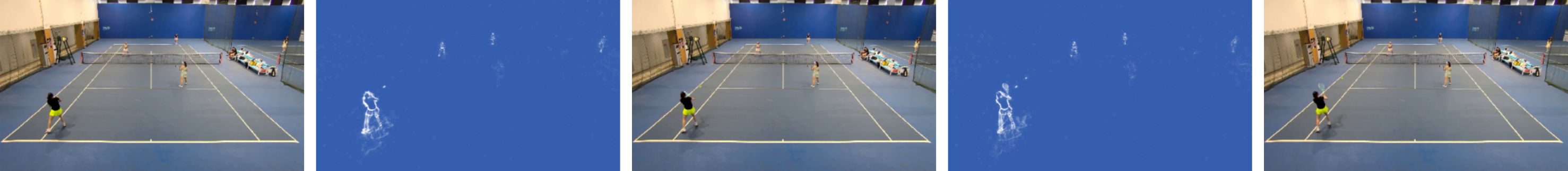}}
\caption{Motion attention map visualizations from the tennis ball tracking~\cite{huang2019tracknet}, shuttlecock tracking~\cite{Sun2020TrackNetV2ES}, and our multi-ball tracking datasets. We select the best model for each dataset to visualize the motion attention maps. Our motion attention maps effectively highlight movements, including those of small objects like balls.}
\label{fig:attn-maps}
\end{figure*}

\begin{figure*}[htbp]
\centering
    % TENNIS
    \subfigure[]
    {\label{fig:pred-tennis-base-1}\includegraphics[trim=0cm 0cm 0cm 0cm, clip=true, width=\linewidth]{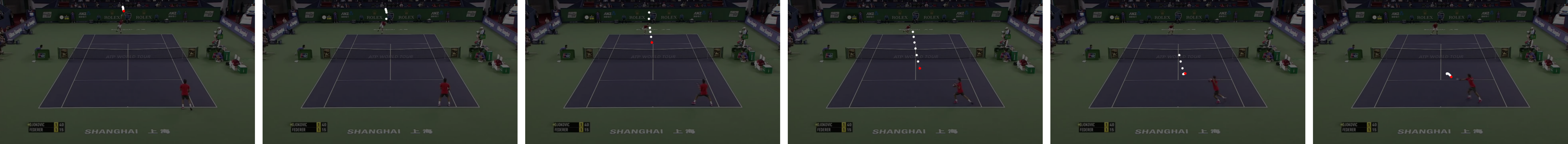}}
    
    \subfigure[]
    {\label{fig:pred-tennis-mars-1}\includegraphics[trim=0cm 0cm 0cm 0cm, clip=true, width=\linewidth]{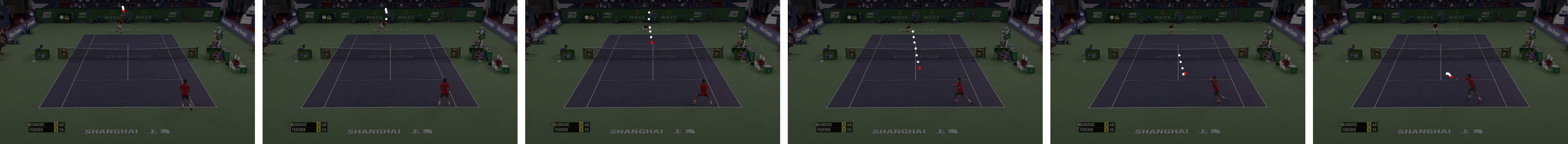}}

    \subfigure[]
    {\label{fig:pred-tennis-base-2}\includegraphics[trim=0cm 0cm 0cm 0cm, clip=true, width=\linewidth]{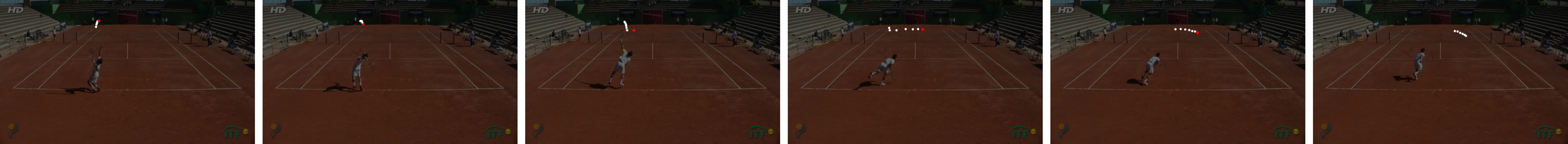}}

    \subfigure[]
    {\label{fig:pred-tennis-mars-2}\includegraphics[trim=0cm 0cm 0cm 0cm, clip=true, width=\linewidth]{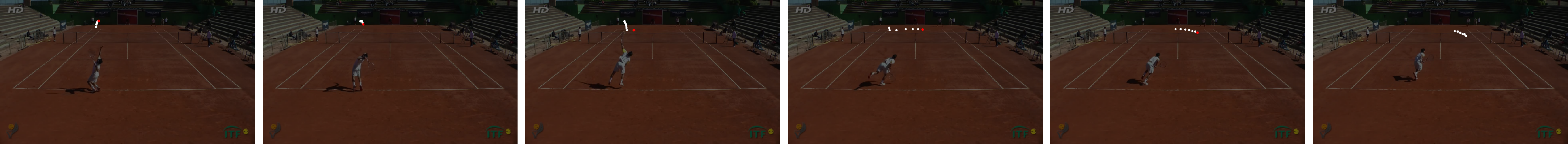}}

    \subfigure[]
    {\label{fig:pred-tennis-base-3}\includegraphics[trim=0cm 0cm 0cm 0cm, clip=true, width=\linewidth]{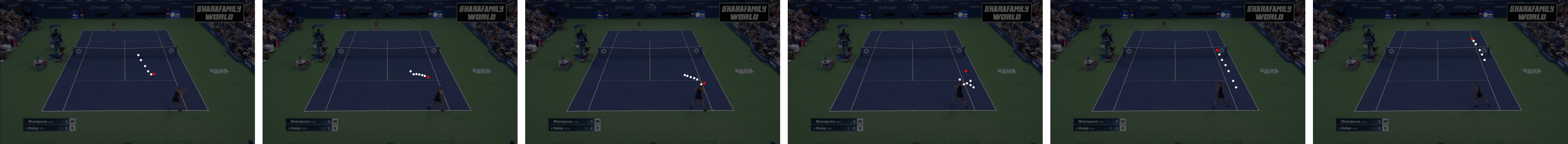}}

    \subfigure[]
    {\label{fig:pred-tennis-mars-3}\includegraphics[trim=0cm 0cm 0cm 0cm, clip=true, width=\linewidth]{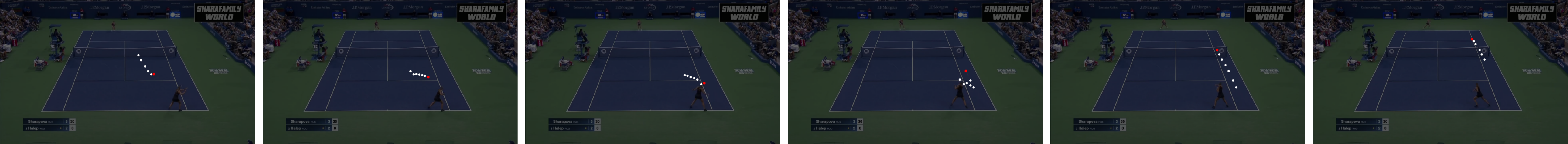}}
\caption{Visualization of ball trajectories on the test set of tennis ball tracking dataset~\cite{huang2019tracknet}. (a), (c) and (e) represent the baseline results, while (b), (d) and (f) show the results from our TrackNetV4. Best viewed with zoom for enhanced detail.}
\label{fig:predictons3}
\end{figure*}

\begin{figure*}[htbp]
\centering
    % BADMINTON
    \subfigure[]
    {\label{fig:pred-badminton-base-1}\includegraphics[trim=0cm 0cm 0cm 0cm, clip=true, width=\linewidth]{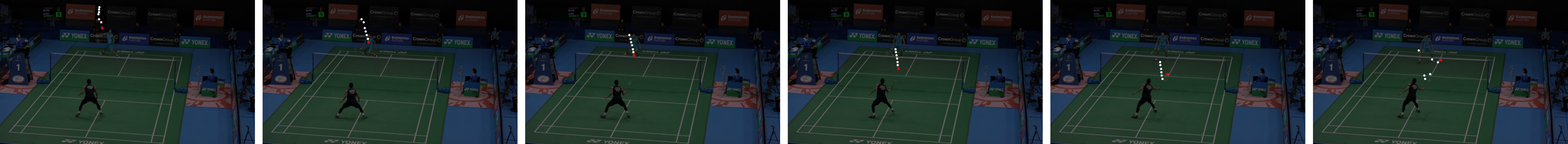}}
    
    \subfigure[]
    {\label{fig:pred-badminton-mars-1}\includegraphics[trim=0cm 0cm 0cm 0cm, clip=true, width=\linewidth]{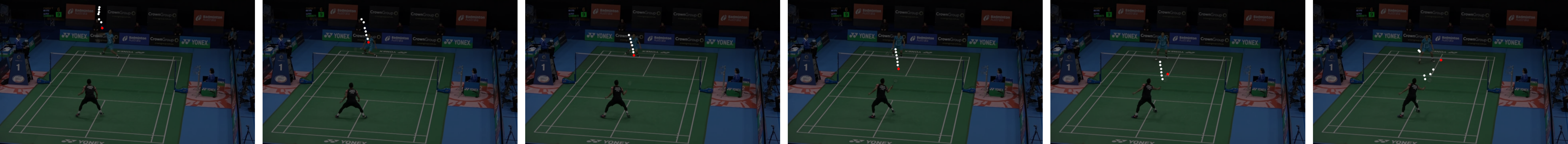}}

    \subfigure[]
    {\label{fig:pred-badminton-base-2}\includegraphics[trim=0cm 0cm 0cm 0cm, clip=true, width=\linewidth]{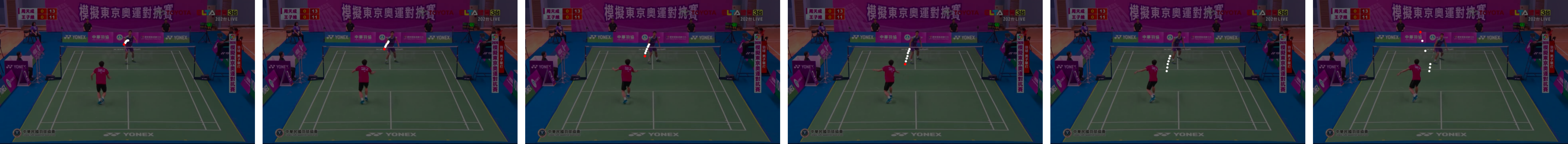}}

    \subfigure[]
    {\label{fig:pred-badminton-mars-2}\includegraphics[trim=0cm 0cm 0cm 0cm, clip=true, width=\linewidth]{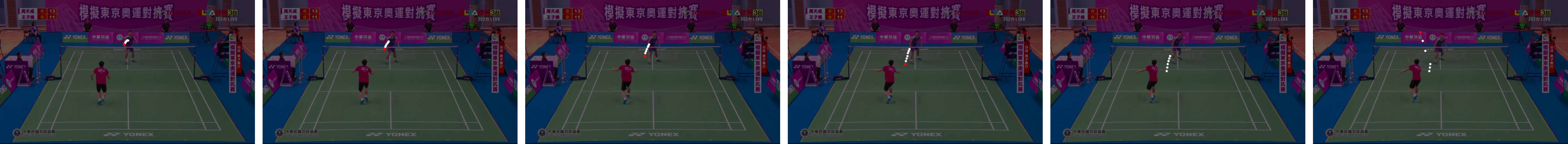}}

    \subfigure[]
    {\label{fig:pred-badminton-base-3}\includegraphics[trim=0cm 0cm 0cm 0cm, clip=true, width=\linewidth]{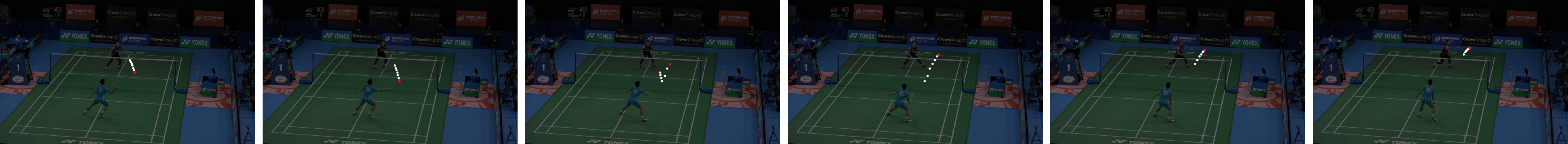}}

    \subfigure[]
    {\label{fig:pred-badminton-mars-3}\includegraphics[trim=0cm 0cm 0cm 0cm, clip=true, width=\linewidth]{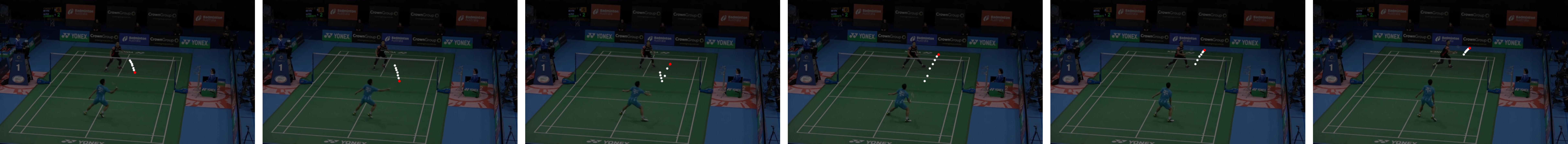}}
\caption{Visualization of ball trajectories on the test set of shuttlecock dataset~\cite{Sun2020TrackNetV2ES}. (a), (c) and (e) represent the baseline results, while (b), (d) and (f) show the results from our TrackNetV4. Best viewed with zoom for enhanced detail.}
\label{fig:predictons2}
\end{figure*}

\begin{figure*}[htbp]
\centering
    \subfigure[]
    {\label{fig:pred-internal-base-1}\includegraphics[trim=0cm 0cm 0cm 0cm, clip=true, width=\linewidth]{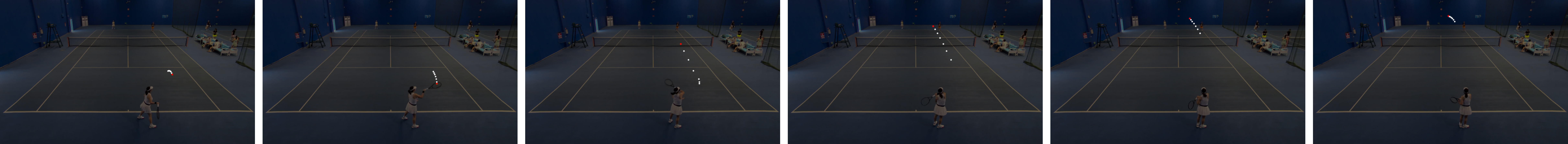}}
    
    \subfigure[]
    {\label{fig:pred-internal-mars-1}\includegraphics[trim=0cm 0cm 0cm 0cm, clip=true, width=\linewidth]{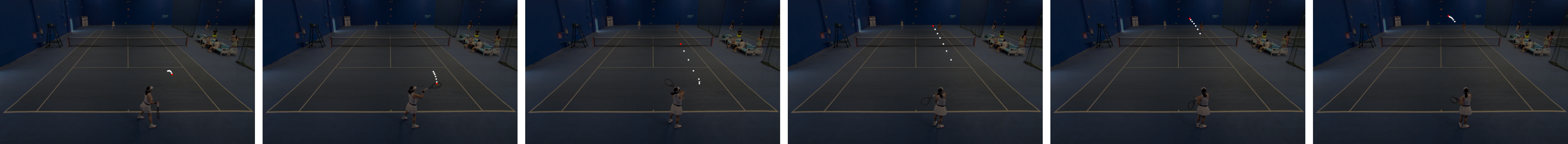}}

    \subfigure[]
    {\label{fig:pred-internal-base-2}\includegraphics[trim=0cm 0cm 0cm 0cm, clip=true, width=\linewidth]{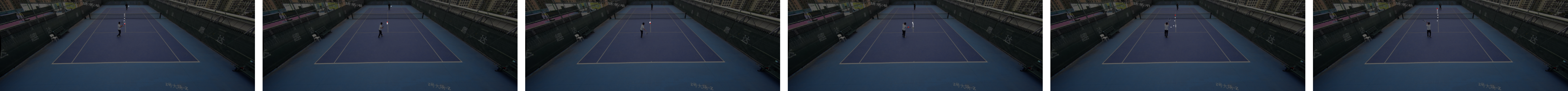}}

    \subfigure[]
    {\label{fig:pred-internal-mars-2}\includegraphics[trim=0cm 0cm 0cm 0cm, clip=true, width=\linewidth]{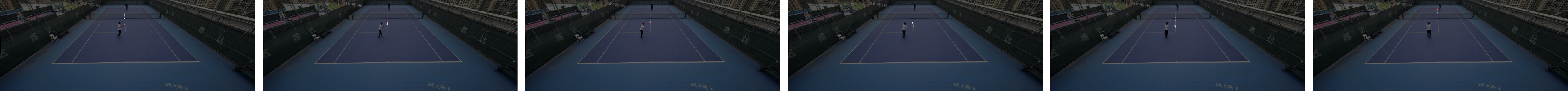}}
\caption{Visualization of ball trajectories on the test set of our multi-ball tracking dataset. (a) and (c) represent the baseline results, while (b) and (d) show the results from our TrackNetV4. Best viewed with zoom for enhanced detail.}
\label{fig:predictons1}
\end{figure*}

\begin{figure*}[htbp]
\centering
    \subfigure[]
    {\label{fig:tennis-1}\includegraphics[trim=0cm 0cm 0cm 0cm, clip=true, width=0.48\linewidth]{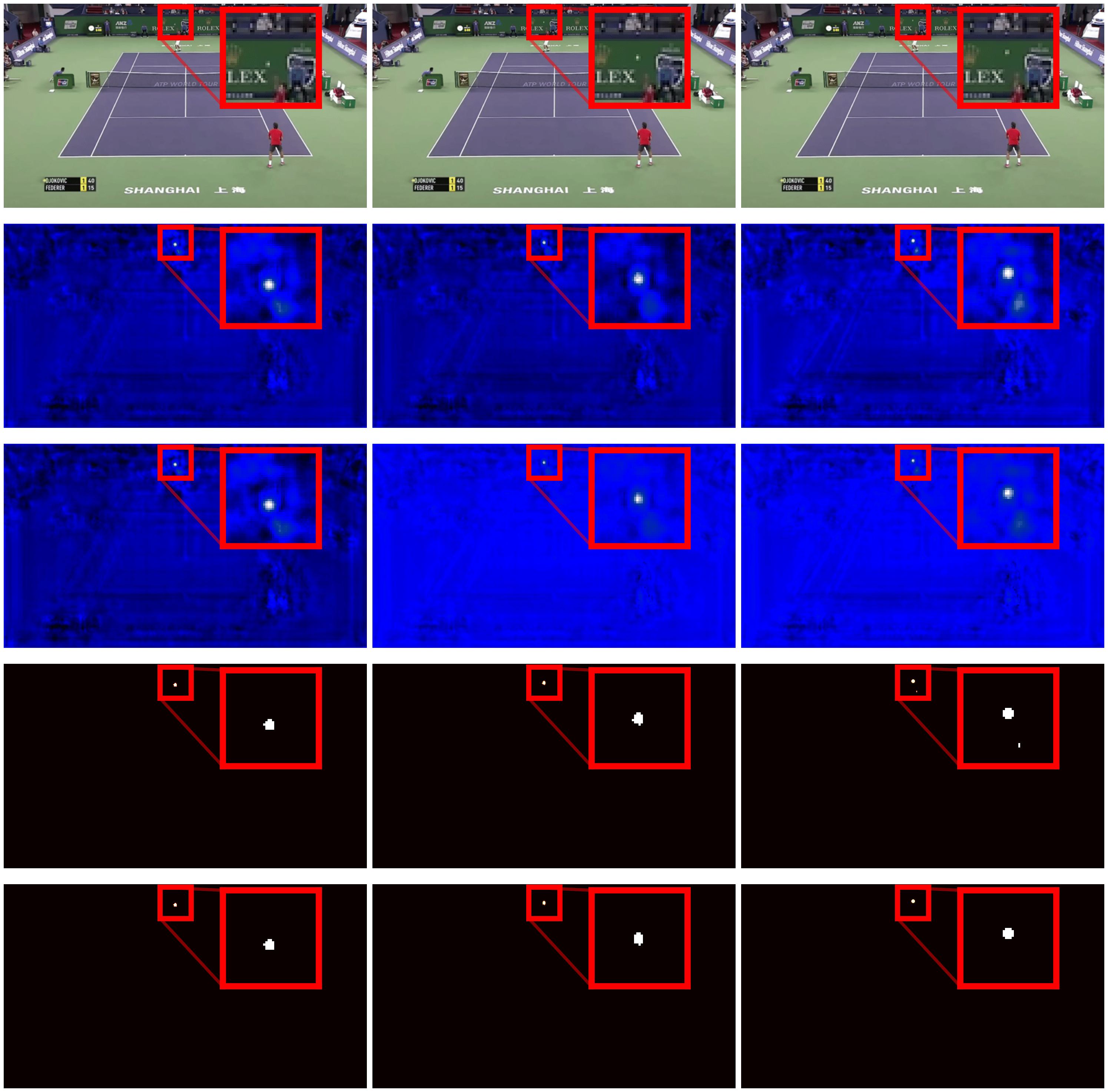}}\hfill
    \subfigure[]
    {\label{fig:tennis-2}\includegraphics[trim=0cm 0cm 0cm 0cm, clip=true, width=0.48\linewidth]{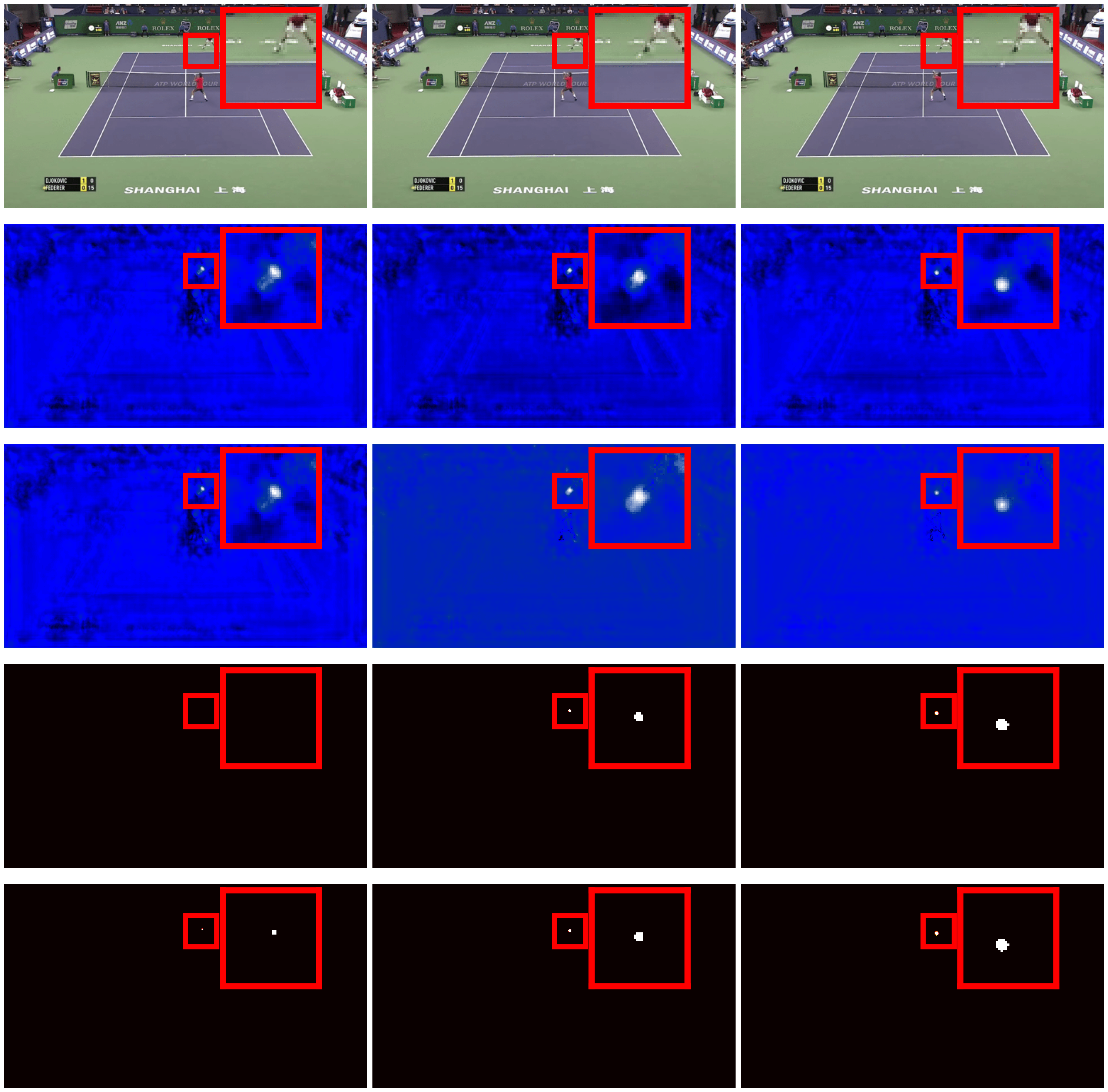}}

    \subfigure[]
    {\label{fig:tennis-3}\includegraphics[trim=0cm 0cm 0cm 0cm, clip=true, width=0.48\linewidth]{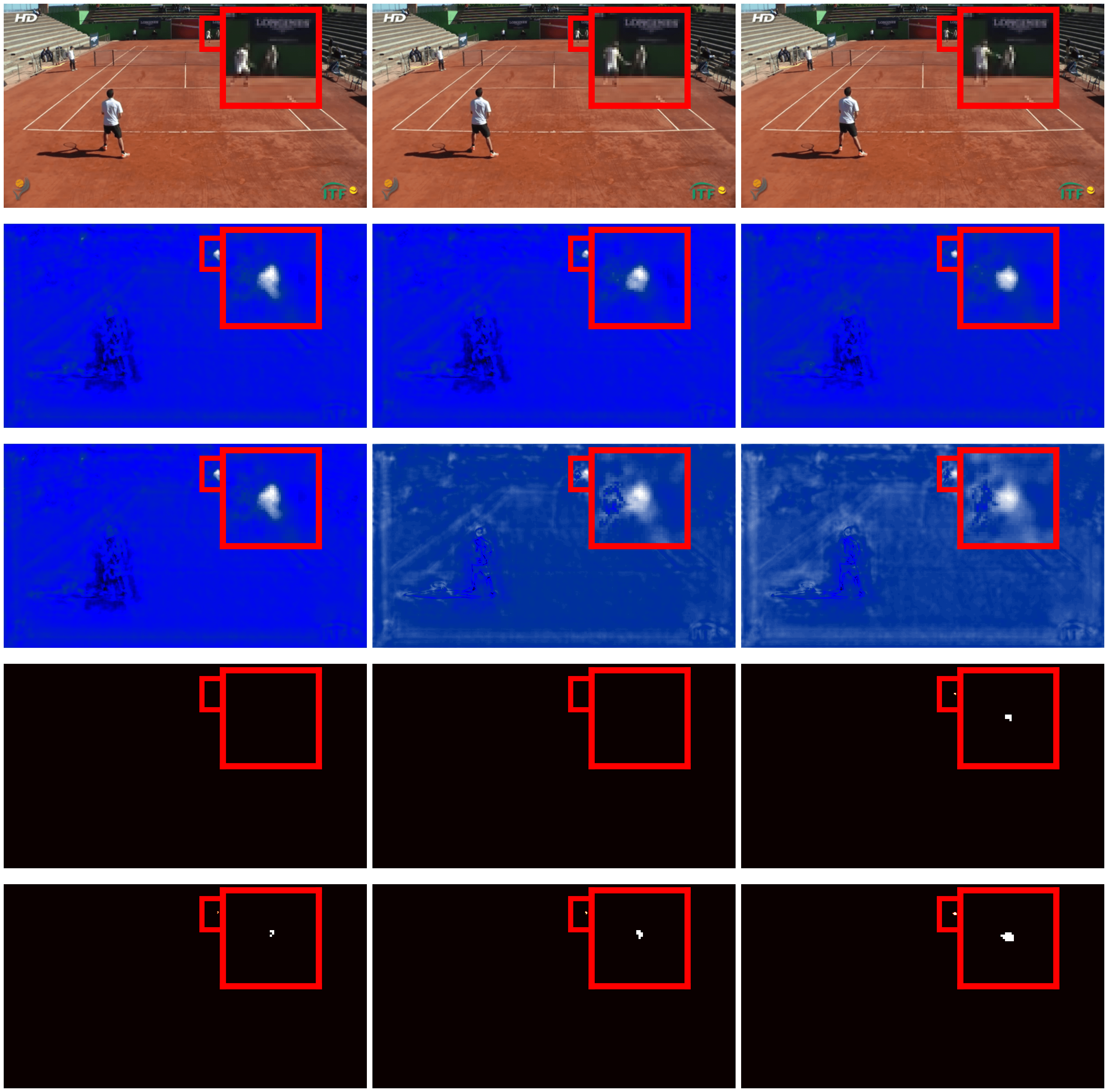}}\hfill
    \subfigure[]
    {\label{fig:tennis-4}\includegraphics[trim=0cm 0cm 0cm 0cm, clip=true, width=0.48\linewidth]{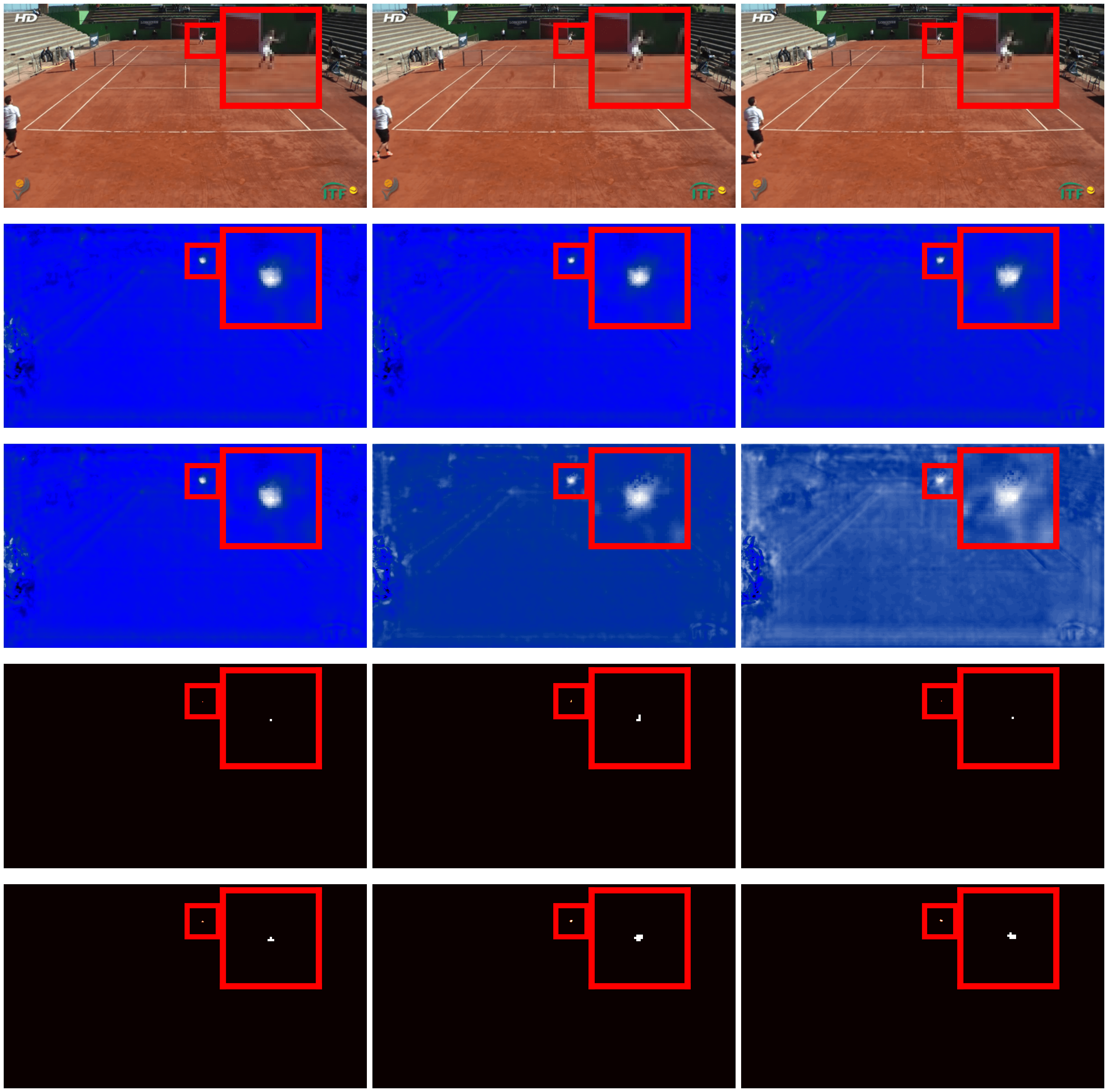}}
    
\caption{Comparison of feature maps and heatmaps with and without motion-aware fusion. We present four groups of visualizations from the tennis tracking dataset~\cite{huang2019tracknet}. For each group, the first row displays the original video frame, the second and third rows show the feature maps from the baseline model and after applying motion-aware fusion, respectively. The fourth and fifth rows present the heatmaps from the baseline model and our TrackNetV4, respectively.}
\label{fig:heat-vis-additional1}
\end{figure*}

\begin{figure*}[htbp]
\centering

    \subfigure[]
    {\label{fig:badminton-1}\includegraphics[trim=0cm 0cm 0cm 0cm, clip=true, width=0.48\linewidth]{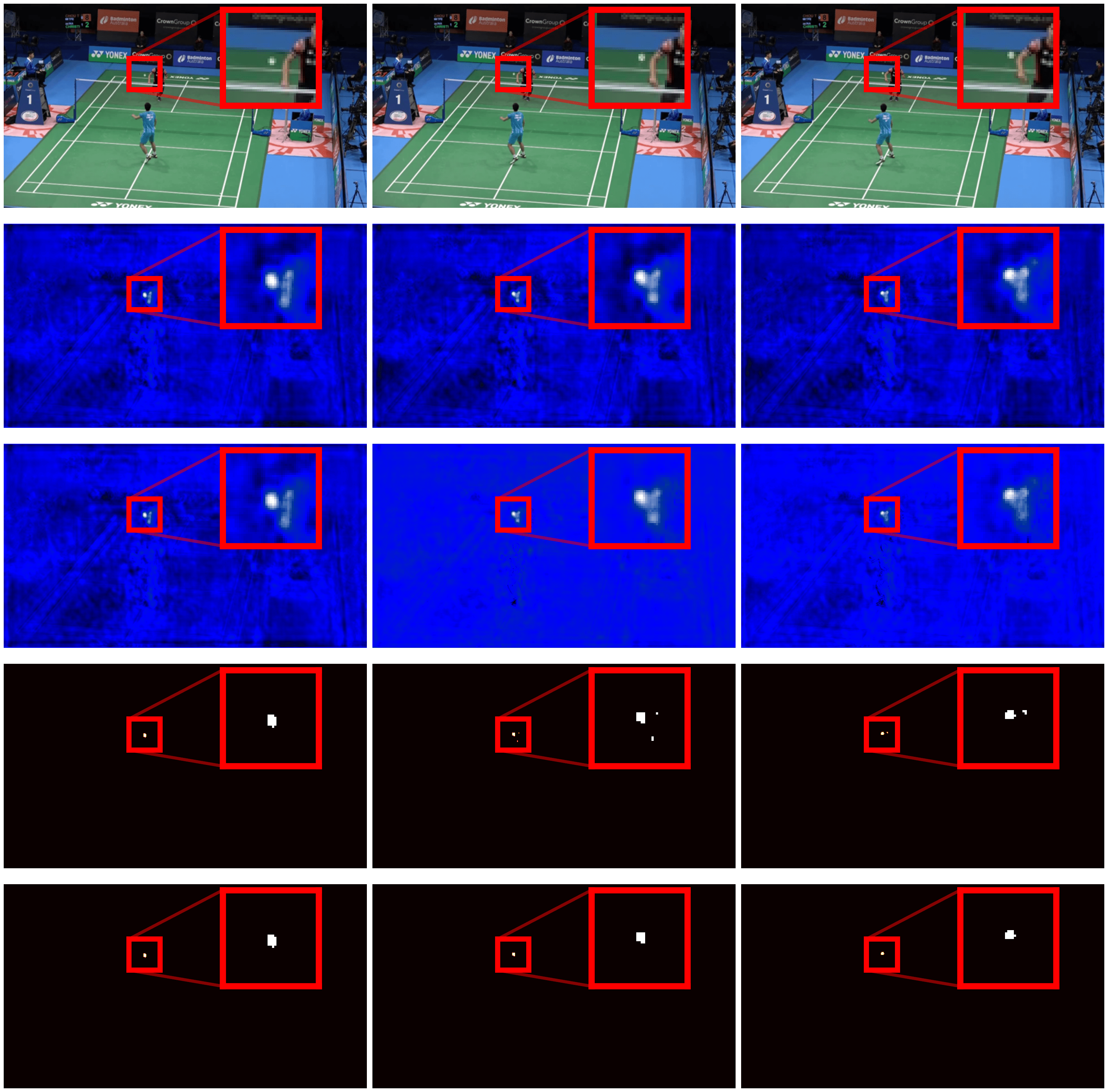}}\hfill
    \subfigure[]
    {\label{fig:badminton-2}\includegraphics[trim=0cm 0cm 0cm 0cm, clip=true, width=0.48\linewidth]{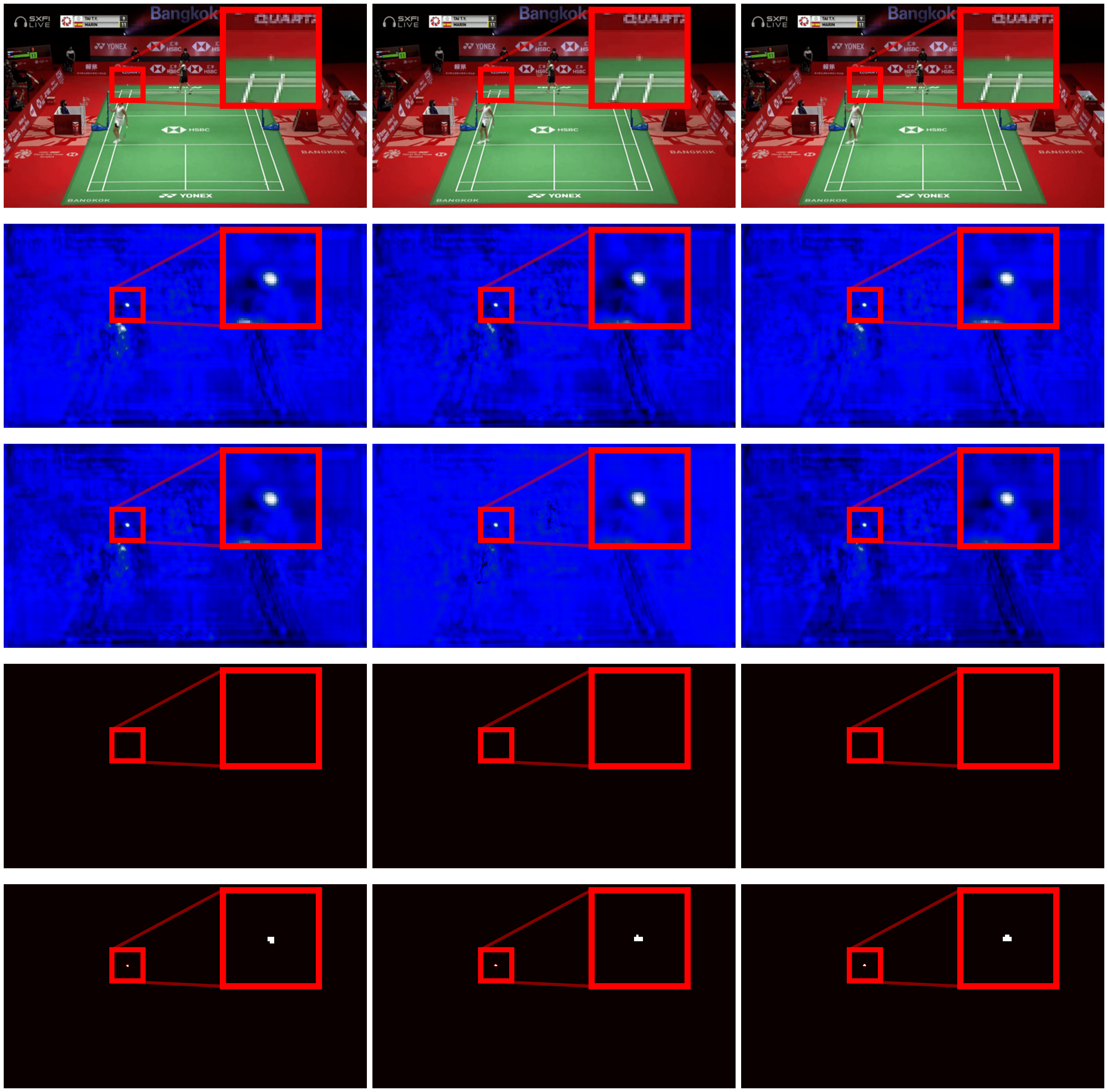}}
    
    \subfigure[]
    {\label{fig:badminton-3}\includegraphics[trim=0cm 0cm 0cm 0cm, clip=true, width=0.48\linewidth]{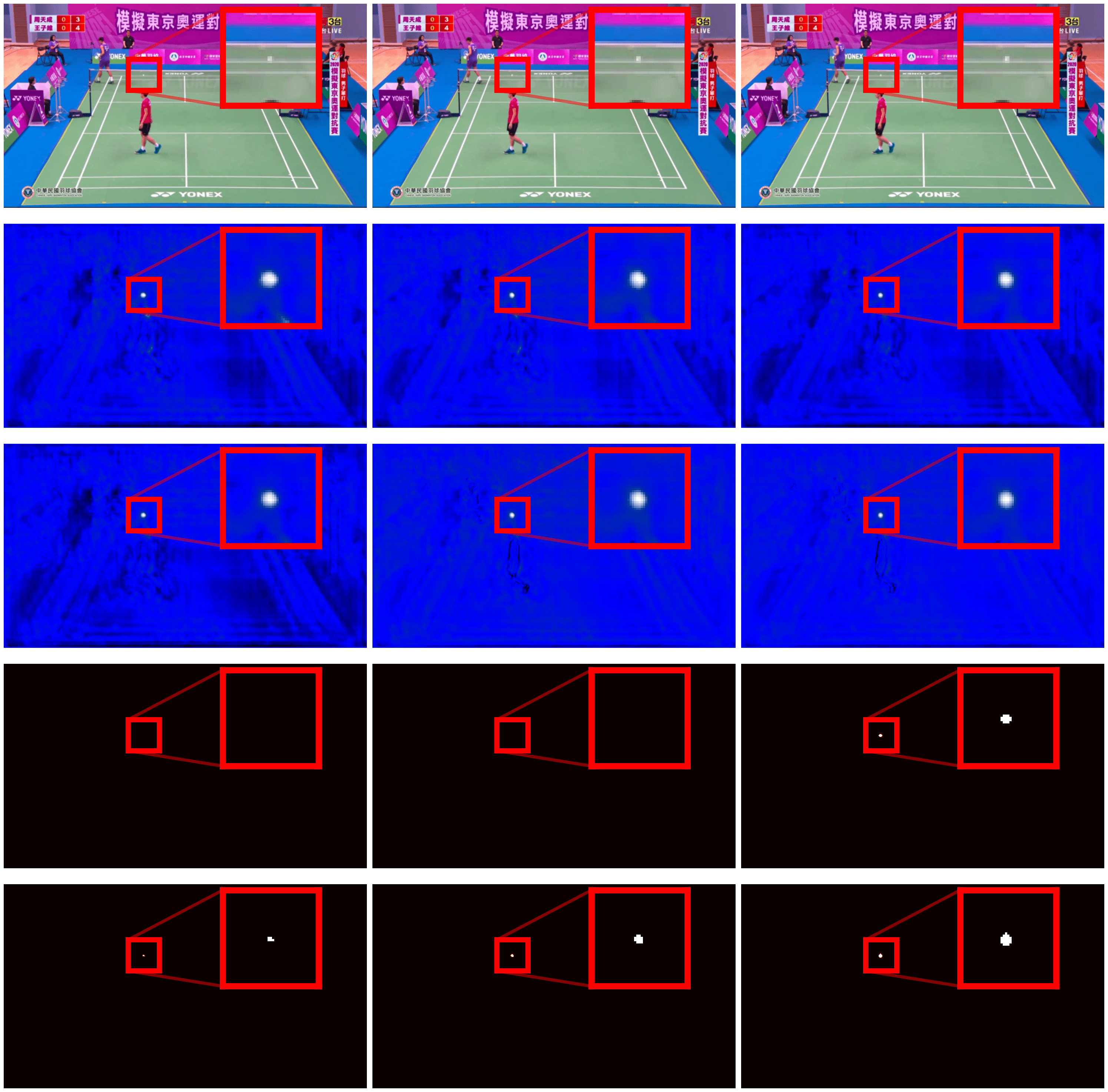}}\hfill
    \subfigure[]
    {\label{fig:badminton-4}\includegraphics[trim=0cm 0cm 0cm 0cm, clip=true, width=0.48\linewidth]{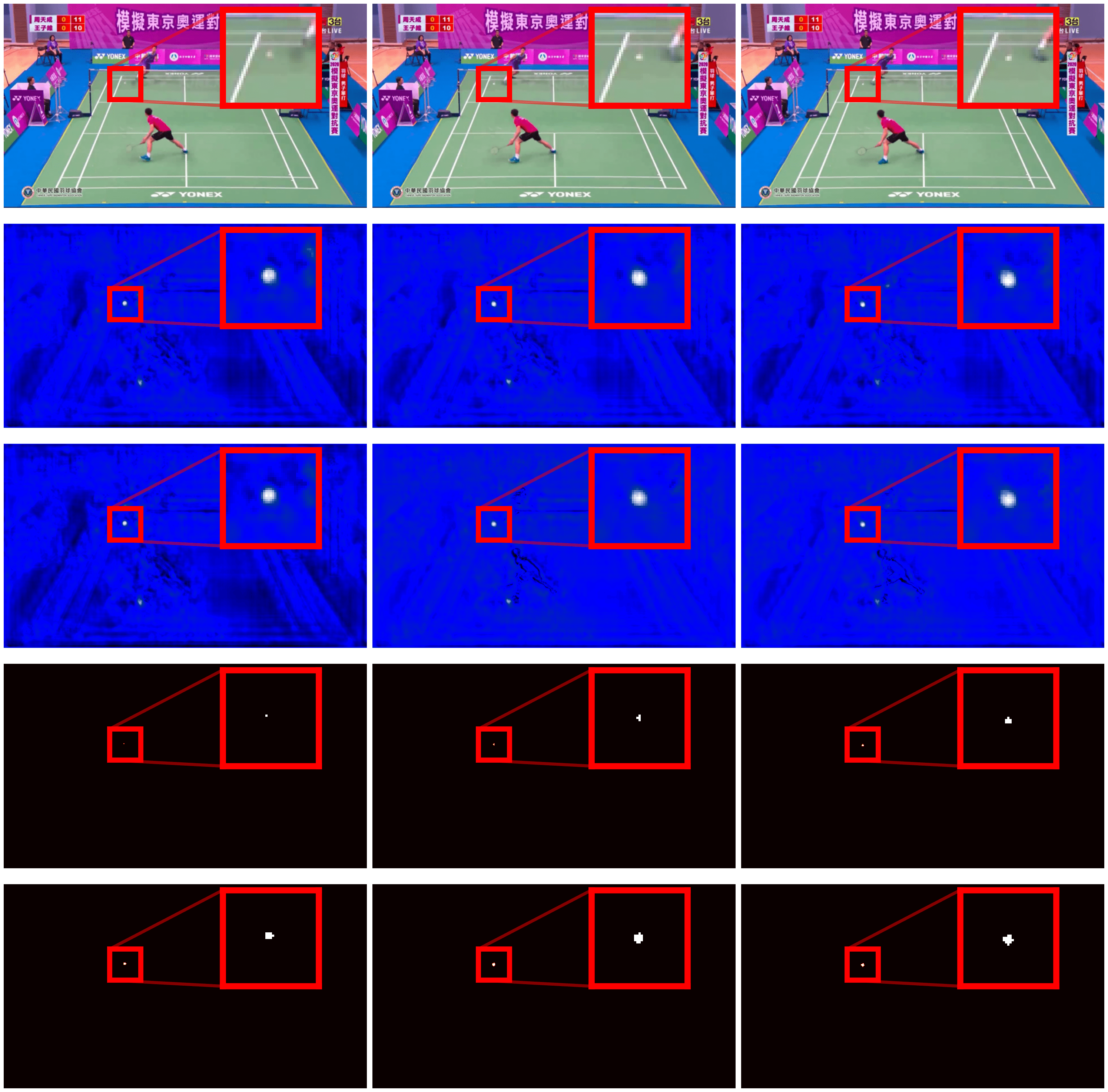}}
    
\caption{Comparison of feature maps and heatmaps with and without motion-aware fusion. We present four groups of visualizations from the shuttlecock dataset~\cite{Sun2020TrackNetV2ES}. For each group, the first row displays the original video frame, the second and third rows show the feature maps from the baseline model and after applying motion-aware fusion, respectively. The fourth and fifth rows present the heatmaps from the baseline model and our TrackNetV4, respectively.}
\label{fig:heat-vis-additional2}
\end{figure*}

\begin{figure*}[htbp]
\centering

    \subfigure[]
    {\label{fig:our-1}\includegraphics[trim=0cm 0cm 0cm 0cm, clip=true, width=0.48\linewidth]{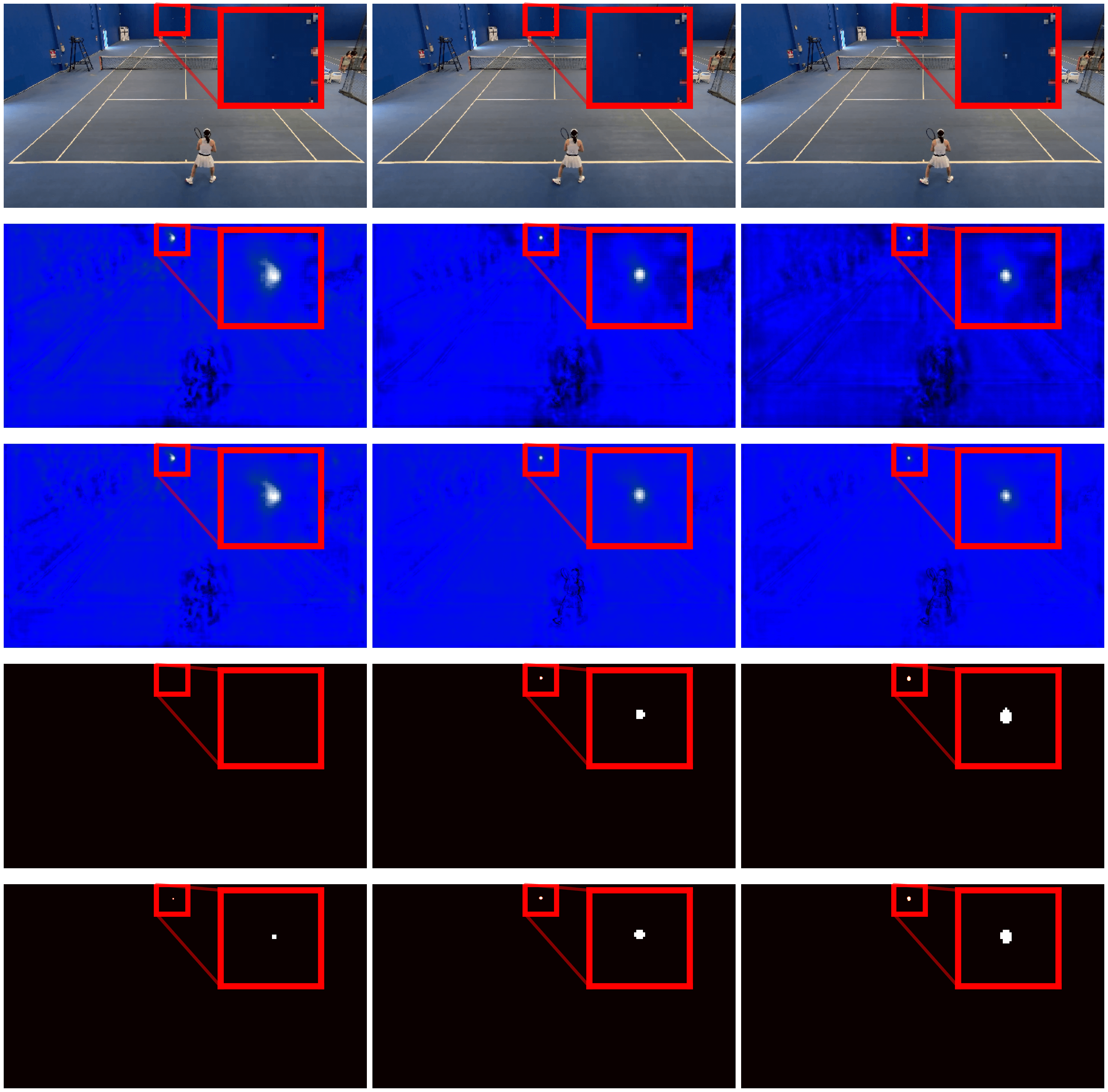}}\hfill
    \subfigure[]
    {\label{fig:our-2}\includegraphics[trim=0cm 0cm 0cm 0cm, clip=true, width=0.48\linewidth]{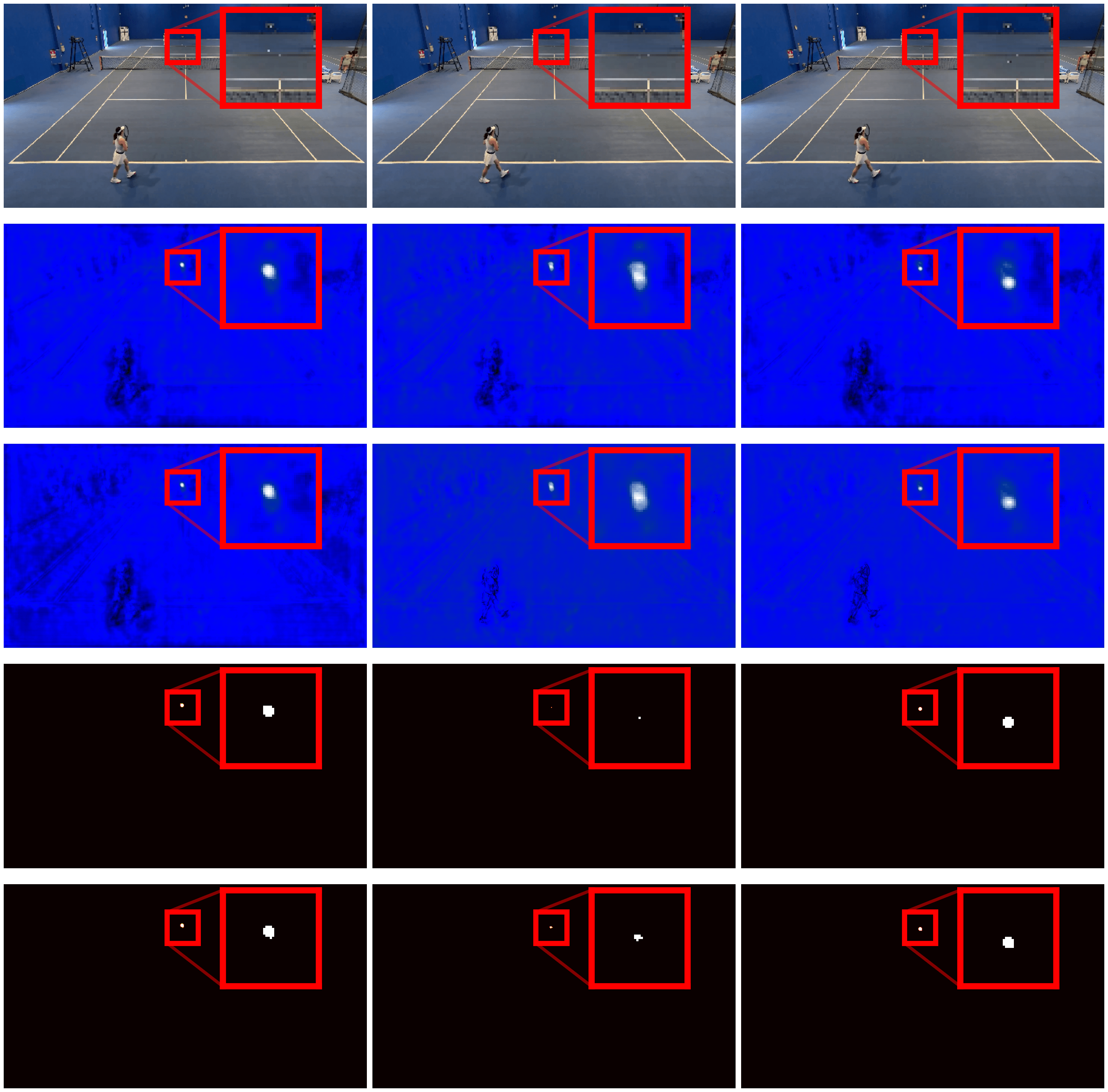}}
    
    \subfigure[]
    {\label{fig:our-3}\includegraphics[trim=0cm 0cm 0cm 0cm, clip=true, width=0.48\linewidth]{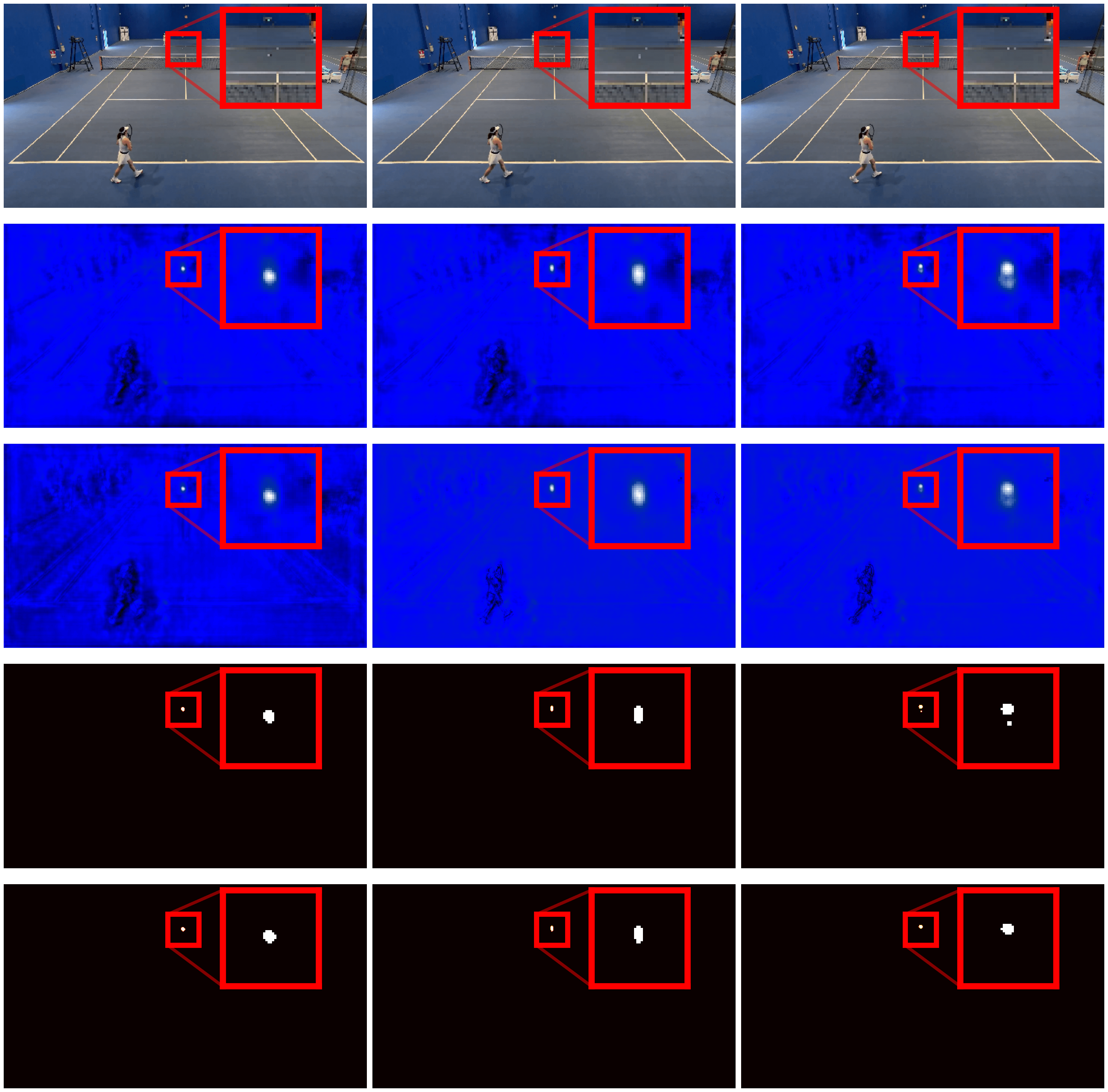}}\hfill
    % \subfigure[]
    % {\label{fig:our-4}\includegraphics[trim=0cm 0cm 0cm 0cm, clip=true, width=0.48\linewidth]{imgs/feature_maps_predictions/1_849.png}}
    % \subfigure[]
    % {\label{fig:our-3}\includegraphics[trim=0cm 0cm 0cm 0cm, clip=true, width=0.48\linewidth]{imgs/feature_maps_predictions/1_917.png}}\hfill
    % \subfigure[]
    {\label{fig:our-4}\includegraphics[trim=0cm 0cm 0cm 0cm, clip=true, width=0.48\linewidth]{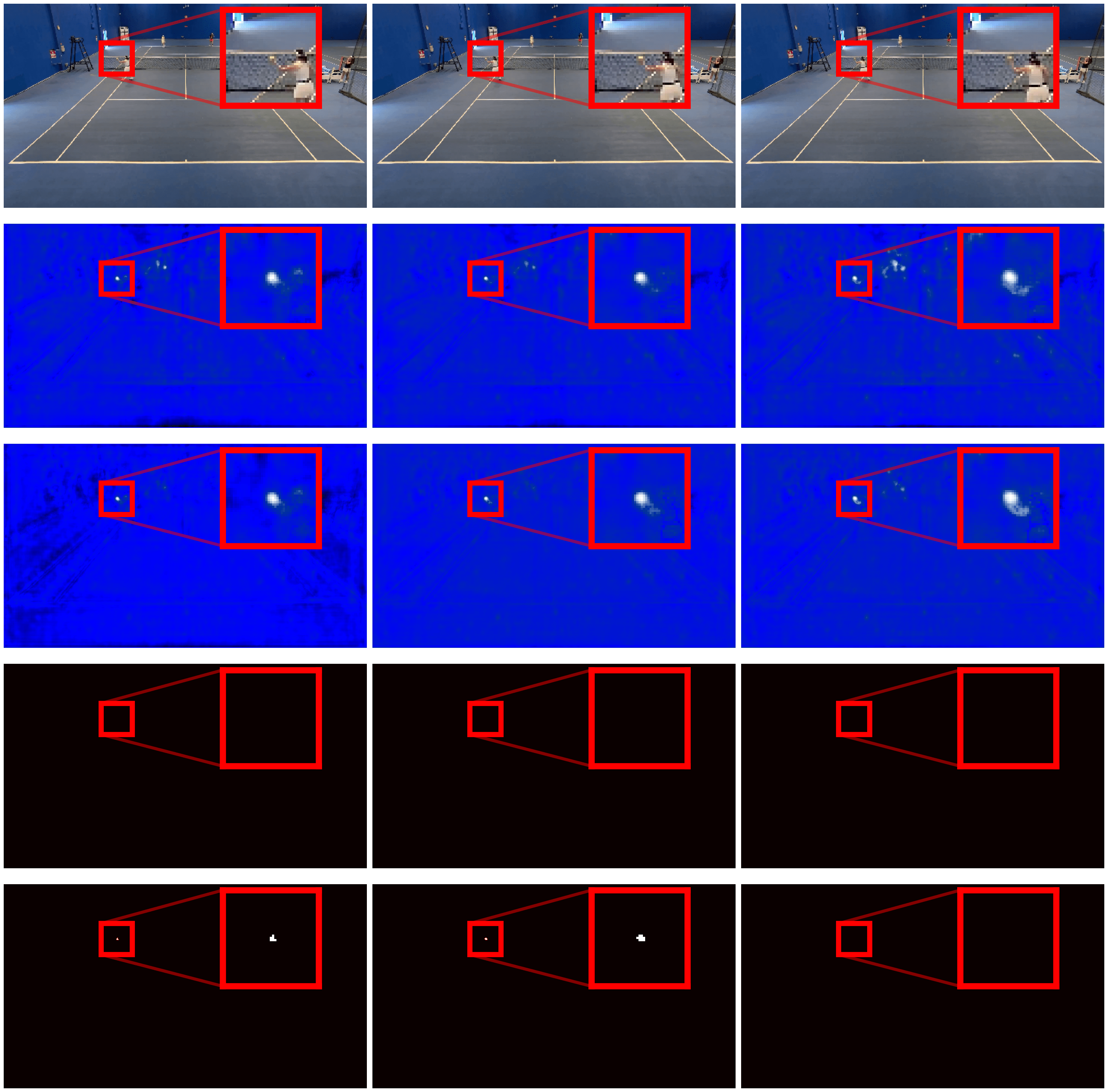}}
    
\caption{Comparison of feature maps and heatmaps with and without motion-aware fusion. We present four groups of visualizations from our multi-ball tracking dataset. For each group, the first row displays the original video frame, the second and third rows show the feature maps from the baseline model and after applying motion-aware fusion, respectively. The fourth and fifth rows present the heatmaps from the baseline model and our TrackNetV4, respectively.}
\label{fig:heat-vis-additional3}
\end{figure*}

\section*{Project Website, Code, and Model}

% \lei{setup our TrackNetV4 website, and update the details here}

Our project website is \href{https://time.anu.edu.au/paper-sites/tracknet-v4}{here}. You can find the code and pre-trained models through the links provided on the site.

\section*{TrackNet Family of Model Architectures}

Below, we provide details of the existing TrackNet frameworks and our TrackNetV4\footnote{We use TrackNetV2 as the backbone for the sake of simplicity.}.

% \lei{Using codes to generate the model architecture and paste it here in table or whatever easy to understand format}

\subsection*{TrackNetV1 introduced in 2019}

TrackNetV1 (Fig.~\ref{fig:tracknetv1}), introduced by Huang \etal in 2019~\cite{huang2019tracknet}, was the first deep learning framework specifically designed for tracking high-speed, small objects in sports applications. The architecture consists of two main components: (i) a VGG-16-based model for object classification, and (ii) a DeconvNet for semantic segmentation. To achieve pixel-level precision in predicting ball locations, the framework employs upsampling techniques to recover information lost in the max-pooling layers. The model uses a symmetric design, with an equal number of upsampling and max-pooling layers, ensuring balanced feature extraction and recovery during the tracking process.

\subsection*{TrackNetV2 introduced in 2020}

TrackNetV2~\cite{Sun2020TrackNetV2ES} is an enhanced version of TrackNetV1, maintaining the same encoder-decoder architecture. The encoder, similar to that of TrackNetV1, leverages VGG-16 to generate feature maps by capturing image clues through convolutional kernels and condensing features via max-pooling operations. The decoder, structured symmetrically to the encoder's downsampling layers, performs upsampling to produce prediction heatmaps with the same resolution as the input images.

TrackNetV2 was designed to improve upon the first-generation model in several key areas, including processing speed, prediction accuracy, and GPU memory efficiency. Key enhancements include a multi-input, multi-output design that allows for more efficient handling of data, the incorporation of skip connections to facilitate better gradient flow and feature preservation, and the introduction of a new weighted Binary Cross-Entropy loss function, which improves performance, particularly in imbalanced data scenarios. These improvements make TrackNetV2 more robust and efficient for high-speed, small-object tracking tasks in sports. Fig.~\ref{fig:tracknetv2} shows the model architecture of TrackNetV2.

\subsection*{TrackNetV3 introduced in 2023}

TrackNetV3~\cite{10.1145/3595916.3626370} is a sophisticated model designed to improve the precision of ball localization in sports applications. It introduces two key modules: (i) Trajectory prediction, which builds on TrackNetV2 by considering a sequence of video frames to generate corresponding heatmaps that indicate the ball’s position over time, and (ii) Trajectory rectification, which refines the predicted trajectory by generating repair masks to assess and correct errors, significantly enhancing the tracking accuracy and completeness of the trajectory.

In addition to these modules, TrackNetV3 uses an estimated background as supplementary data to more precisely locate the ball. Furthermore, inspired by the concept of image inpainting, the model defines an inpainting mask to identify frames that may require correction due to inaccuracies. The trajectory rectification module then takes this mask, along with the predicted trajectory from the tracking module, as inputs to produce a refined, corrected trajectory.

Since the primary focus of this work is the tracking framework, we visualize its architecture in Fig.~\ref{fig:tracknetv3}.

\subsection*{Our TrackNetV4}

Existing tracking frameworks primarily rely on visual features extracted by a VGG-16 network, which are then passed to a DeconvNet model acting as a decoder to predict the pixel-level location of the ball. However, for accurately tracking high-speed, small objects in sports videos, visual features alone are insufficient. These objects require motion information to ensure precise tracking and trajectory prediction.

Inspired by recent advancements in fine-grained video classification tasks~\cite{chen2024motion}, we propose explicitly incorporating motion concepts into the tracking framework. To achieve this, we introduce motion attention maps into the 2D CNN-based TrackNet family -- an architecture that, while traditional, remains highly effective, \eg, TrackNetV2. These motion attention maps are fused with high-level visual feature maps, enhancing motion-sensitive visual representations and improving tracking performance. Fig.~\ref{fig:tracknetv4} shows our model architecture.

Our approach aims to leverage and revitalize existing expert-designed model architectures by integrating modern motion concepts. Through this, we hope to spark renewed interest in these older but still powerful models by demonstrating their potential when combined with contemporary motion-based enhancements.

\section*{Challenging Multi-Ball Tracking Dataset}

A significant issue with existing high-speed, tiny object tracking datasets is their simplicity. These datasets typically feature only a single moving ball on a clean court, leading to relatively straightforward scenarios.

To address this limitation, we collect our own dataset from online sources, incorporating more challenging scenarios, such as multiple moving balls and complex court layouts, which may include more than one court. The aim of our dataset is to demonstrate that TrackNetV4 excels at identifying the primary moving ball in multi-ball scenarios.

% \lei{show some video frames from our dataset, do not choose videos that are able to see the clear faces, or any scenes with chinese characters. The frames should roughly show the following features of our dataset}

Fig.~\ref{fig:internal-dataset} presents visualizations of video frames from our dataset. Our dataset: (i) includes both single- and multi-ball scenarios, with all balls labeled and the primary ball highlighted, (ii) contains videos of varying resolutions, (iii) primarily features amateur games, and (iv) includes both singles and doubles matches. In total, we have collected over 23,000 training frames and more than 1,000 testing frames. Performance and speed metrics are reported.

\section*{Experiments and Discussion}

\begin{table*}[htbp]
\caption{Performance and speed comparisons between the baseline TrackNetV2 and TrackNetV4 (the baseline with our motion-aware fusion) on our dataset are shown. Rows highlighted in blue represent TrackNetV4. Version 1 uses Eq.~\eqref{eq:fusion}, while Version 2 employs the mean motion attention map for element-wise multiplication with each high-level visual feature map. We also report performance results for end-to-end fine-tuning of the pretrained TrackNetV2. The fine-tuning learning rates are set to $1e-3$, $1e-4$, and $1e-5$, respectively, from top to bottom in the finetuning results.}
\begin{center}
\resizebox{\linewidth}{!}{\begin{tabular}{l l c c c c c c c c c c c c}
\toprule
 & \multirow{2}{*}{\bf Method} & \multirow{2}{*}{\bf Total}  & \multicolumn{5}{c}{\bf Confusion matrix} & & \multicolumn{4}{c}{\bf Performance} & \multirow{2}{*}{\bf Speed} \\
\cline{4-8}
\cline{10-13}
& & & TP & TN & FP1 & FP2 & FN & & Acc. & Prec. & Rec. & F1 & \\
\midrule
\multirow{3}{*}{\parbox{2.3cm}{\bf Train from scratch}}
& TrackNetV2 & 3279 & 2999 & 62 & 69 & 7 & 142 &  & 93.4 & 97.5 & 95.5 & 96.5 & 186.4 \\
& TrackNetV4 (version 1) & 3279 & 2971 & 40 & 76 & 29 & 163 &  & 91.8 & 96.6 & 94.8 & 95.7 & 174.7 \\
& TrackNetV4 (version 2) & 3279 & 3047 & 57 & 29 & 12 & 134 &  & \textbf{94.7} & \textbf{98.7} & \textbf{95.8} & \textbf{97.2} & 169.1 \\
\hline
\multirow{6}{*}{\parbox{2.3cm}{\bf Fine-tuning}}
& TrackNetV4 (version 1) & 3279 & 2982 & 61  & 47 & 8 & 181 &  & 92.8 & \textbf{98.2} & 94.3 & 96.2 & 161.1 \\
& TrackNetV4 (version 2) & 3279 & 3050 & 60 & 44 & 9 & 116 &  & \textbf{94.8} & \textbf{98.3} & \textbf{96.3} & \textbf{97.3} & 168.5 \\
& TrackNetV4 (version 1) & 3279 & 3030 & 61 & 59 & 8 & 121 &  & \textbf{94.3} & \textbf{97.8} & \textbf{96.2} & \textbf{97.0} & 169.3 \\
& TrackNetV4 (version 2) & 3279 & 2721 & 63 & 35 & 6 & 454 &  & 84.9 & \textbf{98.5} & 85.7 & 91.7 & 171.3 \\
& TrackNetV4 (version 1) & 3279 & 2914 & 61 & 51 & 8 & 245 &  & 90.7 & \textbf{98.0} & 92.2 & 95.0 & 173.0 \\
& TrackNetV4 (version 2) & 3279 & 3022 & 54 & 94 & 15 & 94 &  & \textbf{93.8} & 96.5 & \textbf{97.0} & \textbf{97.0} & 179.2 \\
\bottomrule
\end{tabular}}
\label{tab:ours}
\end{center}
\end{table*}

We use TrackNetV2 as a baseline to demonstrate that, when enhanced with our motion attention maps and motion-aware fusion, referred to as TrackNetV4, it performs effectively on our challenging tennis ball tracking dataset. For all experiments, we adhere to the parameter settings specified by the original authors for training models from scratch, including setting the number of training epochs to 30 and the initial learning rate to 1.0.

We select two fusion variants: version 1 follows Eq.~\eqref{eq:fusion}, while version 2 uses the mean motion attention map for element-wise multiplication. We also present the results of end-to-end fine-tuning of TrackNetV4, starting from the pretrained model, using different learning rates. Table~\ref{tab:ours} summarizes the results.

As shown in the table, TrackNetV4 (version 2) trained from scratch outperforms the baseline, with improvements of 1.3\%, 1.2\%, 0.3\%, and 0.7\% in accuracy, precision, recall, and F1-score, respectively. Additionally, we observe that fine-tuning on top of the pretrained baseline further boosts performance, particularly in the recall metric.

% \lei{document both train from scratch and finetuning with different learning rates experimental results to the table.}

\section*{The Role of Fine-Tuning}

As shown in Table~\ref{tab:ours}, fine-tuning our TrackNetV4 using a pretrained TrackNetV2 model on our dataset generally improves performance, particularly for the version 2 model. For instance, with a learning rate of $1e-3$ and two additional learnable parameters (TrackNetV4 version 2) on top of the TrackNetV2 pretrained baseline, performance increased by 1.4\%, 0.8\%, 0.8\%, and 0.8\% for accuracy, precision, recall, and F1-score, respectively. Even with version 1 of our TrackNetV4, we observed improvements over the TrackNetV2 baseline of 0.9\%, 0.3\%, 0.7\%, and 0.5\% for accuracy, precision, recall, and F1-score, respectively. This demonstrates that both our motion attention maps and motion-aware fusion mechanism play crucial roles in enhancing the network's learning capacity for motion concepts.

% \lei{show the weights comparison using cosine similarity, as in Qixiang's paper. Just using pretrained model (baseline) and the one finetuned, for TrackNetV3.}

On the other hand, training from scratch provides a similar boost across all four performance metrics.
Fig.~\ref{fig:similarity_plot} shows the per-layer weight similarity for both training from scratch and fine-tuning the baseline using TrackNetV4.
We observe significant differences in the per-layer weights between fine-tuning with the pretrained TrackNetV2 (using a smaller learning rate, such as $1e-3$) and training from scratch with TrackNetV4 (starting with a learning rate of 1.0), particularly in the 2D convolutional layers.
We believe the motion prompt layer plays a crucial role in guiding the network to extract motion-aware features for tracking and prediction tasks. In the model trained from scratch, motion provides fresh information to the 2D CNN tracking framework. On the other hand, injecting motion attention maps into the pretrained model, even when using a fine-tuning strategy, still enhances tracking performance.

\section*{Visualizations of Motion Attention Maps}

Fig.~\ref{fig:attn-maps} shows visualizations of the learned motion attention maps generated by the best model for each dataset.

As illustrated, these motion attention maps effectively capture and highlight the motion dynamics within the videos, including small objects like balls.
This ability to focus on fine details demonstrates the model's capacity to accurately represent subtle movements, enhancing its overall performance in high-speed and small-object tracking tasks.

% \lei{show some motion attention maps for the visualisations. Choose the best model, get the frame differencing maps, take absolute value and using the learned a and b, to modulate it then visualise the attention maps. The visualisation should cover (i) original 3-frame, 2 motion attention maps for these 3.}

\section*{Visualization of Ball Trajectories}

Fig.~\ref{fig:predictons3},~\ref{fig:predictons2} and~\ref{fig:predictons1} present visualizations of predicted ball locations in the form of trajectories.

As shown in these figures, we observe that (i) the trajectories from both the baseline model (TrackNetV2) and our TrackNetV4, which builds on top of the baseline, are very similar for both the tennis ball and shuttlecock tracking datasets, and (ii) our TrackNetV4 on our multi-ball tracking dataset, using motion attention maps and motion-aware fusion, generates significantly smoother and more accurate ball trajectories. This highlights the effectiveness of incorporating motion concepts for accurately tracking and predicting fast-moving, small objects.

% \lei{show some trajectory prediction results from 3 datasets (including our own dataset), show videos in the form of continuous video frames, 6 per raw, choose challenging ones, show that ours are better and more accurate than baseline, this should form eg 8 x 6, eg 3 for badminton, 3 for tennis, and 2 for ours, each set show baseline versus ours, resulting in 16x6 in total. For the selected 6 frames, you might need to skip few frames to make it look nice and more like a changing video frame style.}

\section*{Additional Visualizations}

Below in Fig \ref{fig:heat-vis-additional1}, ~\ref{fig:heat-vis-additional2} and~\ref{fig:heat-vis-additional3}, we present additional visualizations comparing feature maps and heatmaps with and without our motion-aware fusion.

% As shown in these figures, despite the minor visual differences, we observe that our motion-aware fusion produces precise and clear ball locations, even when they are small, blurry, exhibit afterimage trails, or are nearly invisible.

We observe that the motion-aware fusion significantly enhances the tracking and prediction of ball locations, as demonstrated by the clearer visualizations of both the feature maps (before applying the Sigmoid function) and the generated heatmaps. Notably, our feature maps show greater clarity, and the resulting heatmaps exhibit enhanced robustness, even when the balls are small, blurry, have afterimage trails, or are nearly invisible. This highlights the effectiveness of our method, which remains lightweight while successfully tracking high-speed, small objects in sports scenarios.

% \lei{prepare more visualisations like Fig. 4 in the main paper: original video frames, feature maps and heatmaps. Fig. 1 shows 1 set, in appendix, we want to show \eg, 10 sets. Choose different and diverse scenarios, with both badminton and tennis balls scenarios.}

\section*{Questions and Answers}

% \lei{Please rewrite the answers and make it match with your thoughts and make those questions and answers indeed useful for future researchers. We need to expand some of the answers too by providing more detailed plans}

\textbf{Q1: What motivates you for this research project?}

\textbf{A1:} AI has powered numerous real-world applications, including smart coaching apps for sports analysis and player performance monitoring. One of our key motivations is to improve tracking and prediction accuracy for fast-moving, small objects in sports videos. Traditional tracking methods often struggle with high-speed objects and frequent occlusions, leading to inaccurate predictions and poor performance. By enhancing tracking capabilities with motion-based concepts, we aim to provide more reliable tools for sports analysis and other applications requiring precise object tracking.

Current TrackNet models rely on traditional 2D CNNs for video-based ball tracking. While these models are effective, there is certainly room for improvement. Our goal is to revisit these well-established, expert-designed architectures and enhance them by integrating modern motion prompts as guidance to boost ball tracking and prediction performance. We aim to reignite interest in reusing existing model architectures and pretrained models to reduce energy consumption and computing resource demands. We are the first to propose the concept of environment-friendly and reusable strategies in the fields of computer vision and machine learning.

\textbf{Q2: What are the core innovations of TrackNetV4 and the reason for its name?}

\textbf{A2:} TrackNetV4 builds on existing TrackNet models to enhance the tracking and prediction of high-speed, small objects in sports activities, using the latest motion attention maps and motion-aware fusion mechanisms. Using TrackNetV2 as a foundation, we introduce a motion prompt layer~\cite{chen2024motion} that generates a sequence of motion attention maps. These maps are then fused with high-level visual feature maps through element-wise multiplication. Despite its simplicity, requiring only two additional learnable parameters, this fusion consistently improves accuracy, precision, recall, and F1-score.

We are the first to incorporate motion concepts into traditional 2D CNN-based tracking frameworks. TrackNetV4, when built on TrackNetV3 as the baseline architecture and augmented with motion attention maps and the fusion mechanism, outperforms TrackNetV3. These innovations lead to improved tracking performance and robustness, particularly in challenging scenarios with fast-moving objects and occlusions.

The core innovations of TrackNetV4 lie in its enhanced motion dynamics modeling, advanced fusion techniques, and improved feature extraction mechanisms. The name `TrackNetV4' marks its evolution from earlier versions, representing the fourth iteration with substantial improvements and new capabilities. Furthermore, by reusing established tracking architectures and integrating modern modules, TrackNetV4 establishes itself as a next-generation framework with measurable improvements across all performance metrics.

\textbf{Q3: Would the new challenging multi-ball object tracking dataset be released in the future?}

\textbf{A3:} Yes, the release of a new challenging multi-ball object tracking dataset is planned for the future. This dataset aims to address current limitations by providing more complex and diverse scenarios for object tracking, which will facilitate the development and evaluation of more robust tracking algorithms.

Our project website includes visualizations of ball tracking performance on the test set of this dataset. Additionally, we provide access to our model code, best-performing models, and the inference, testing, and evaluation scripts for interested researchers.

\textbf{Q4: How does TrackNetV4 advance object tracking and model reuse?}

\textbf{A4:} TrackNetV4 has a range of potential applications, including real-time sports analysis, video surveillance, autonomous vehicles, and robotics. Its enhanced tracking capabilities make it suitable for scenarios requiring precise and reliable object tracking, even in dynamic and cluttered environments.

TrackNetV4 represents a pioneering step towards reusable models and environmentally-friendly training concepts. We aim to inspire renewed interest in integrating modern, innovative modules into existing expert-designed, pretrained models with minimal modifications. This approach, from a practical perspective, seeks to benefit the research community by enhancing the utility of established models.

{\small
\bibliographystyle{plain}
\bibliography{egbib}

\begin{thebibliography}{10}

\bibitem{chen2024motion}
Qixiang Chen, Lei Wang, Piotr Koniusz, and Tom Gedeon.
\newblock Motion meets attention: Video motion prompts.
\newblock {\em Asian Conference on Machine Learning (ACML)}, 2024.

\bibitem{chen2024sato}
Wenshuo Chen, Hongru Xiao, Erhang Zhang, Lijie Hu, Lei Wang, Mengyuan Liu, and
  Chen Chen.
\newblock Sato: Stable text-to-motion framework.
\newblock {\em ACM Multimedia (ACM-MM)}, 2024.

\bibitem{10.1145/3595916.3626370}
Yu-Jou Chen and Yu-Shuen Wang.
\newblock Tracknetv3: Enhancing shuttlecock tracking with augmentations and
  trajectory rectification.
\newblock In {\em Proceedings of the 5th ACM International Conference on
  Multimedia in Asia}, MMAsia '23, New York, NY, USA, 2024. Association for
  Computing Machinery.

\bibitem{duan2024cross}
Haoyi Duan, Yan Xia, Zhou Mingze, Li~Tang, Jieming Zhu, and Zhou Zhao.
\newblock Cross-modal prompts: Adapting large pre-trained models for
  audio-visual downstream tasks.
\newblock {\em Advances in Neural Information Processing Systems}, 36, 2024.

\bibitem{fang2024signllm}
Sen Fang, Lei Wang, Ce~Zheng, Yapeng Tian, and Chen Chen.
\newblock Signllm: Sign languages production large language models.
\newblock {\em arXiv preprint arXiv:2405.10718}, 2024.

\bibitem{huang2019tracknet}
Yu-Chuan Huang, I-No Liao, Ching-Hsuan Chen, Ts{\`\i}-U{\'\i} {\.I}k, and
  Wen-Chih Peng.
\newblock Tracknet: A deep learning network for tracking high-speed and tiny
  objects in sports applications.
\newblock In {\em 2019 16th IEEE International Conference on Advanced Video and
  Signal Based Surveillance (AVSS)}, pages 1--8. IEEE, 2019.

\bibitem{NEURIPS2023_656678aa}
Bing Li, Jiaxin Chen, Xiuguo Bao, and Di~Huang.
\newblock Compressed video prompt tuning.
\newblock In A.~Oh, T.~Naumann, A.~Globerson, K.~Saenko, M.~Hardt, and
  S.~Levine, editors, {\em Advances in Neural Information Processing Systems},
  volume~36, pages 31895--31907. Curran Associates, Inc., 2023.

\bibitem{noh2015learning}
Hyeonwoo Noh, Seunghoon Hong, and Bohyung Han.
\newblock Learning deconvolution network for semantic segmentation.
\newblock In {\em Proceedings of the IEEE international conference on computer
  vision}, pages 1520--1528, 2015.

\bibitem{simonyan2014very}
Karen Simonyan and Andrew Zisserman.
\newblock Very deep convolutional networks for large-scale image recognition.
\newblock {\em arXiv preprint arXiv:1409.1556}, 2014.

\bibitem{Sun2020TrackNetV2ES}
Nien-En Sun, Yu-Ching Lin, Shao-Ping Chuang, Tzu-Han Hsu, Dung-Ru Yu, Ho-Yi
  Chung, and Ts{\`i}-U{\'i} İk.
\newblock Tracknetv2: Efficient shuttlecock tracking network.
\newblock {\em 2020 International Conference on Pervasive Artificial
  Intelligence (ICPAI)}, pages 86--91, 2020.

\bibitem{wang2023yolov7}
Chien-Yao Wang, Alexey Bochkovskiy, and Hong-Yuan~Mark Liao.
\newblock Yolov7: Trainable bag-of-freebies sets new state-of-the-art for
  real-time object detectors.
\newblock In {\em Proceedings of the IEEE/CVF conference on computer vision and
  pattern recognition}, pages 7464--7475, 2023.

\bibitem{wang2024vilt}
Hao Wang, Fang Liu, Licheng Jiao, Jiahao Wang, Zehua Hao, Shuo Li, Lingling Li,
  Puhua Chen, and Xu~Liu.
\newblock Vilt-clip: Video and language tuning clip with multimodal prompt
  learning and scenario-guided optimization.
\newblock In {\em Proceedings of the AAAI Conference on Artificial
  Intelligence}, volume~38, pages 5390--5400, 2024.

\bibitem{wang2024flow}
Lei Wang and Piotr Koniusz.
\newblock Flow dynamics correction for action recognition.
\newblock In {\em ICASSP 2024-2024 IEEE International Conference on Acoustics,
  Speech and Signal Processing (ICASSP)}, pages 3795--3799. IEEE, 2024.

\bibitem{wang2024high}
Lei Wang, Ke~Sun, and Piotr Koniusz.
\newblock High-order tensor pooling with attention for action recognition.
\newblock In {\em ICASSP 2024-2024 IEEE International Conference on Acoustics,
  Speech and Signal Processing (ICASSP)}, pages 3885--3889. IEEE, 2024.

\bibitem{wang2024taylor}
Lei Wang, Xiuyuan Yuan, Tom Gedeon, and Liang Zheng.
\newblock Taylor videos for action recognition.
\newblock {\em International Conference on Machine Learning (ICML)}, 2024.

\bibitem{xu2024finepose}
Jinglin Xu, Yijie Guo, and Yuxin Peng.
\newblock Finepose: Fine-grained prompt-driven 3d human pose estimation via
  diffusion models.
\newblock In {\em Proceedings of the IEEE/CVF Conference on Computer Vision and
  Pattern Recognition}, pages 561--570, 2024.

\end{thebibliography}
}

% \pagebreak

\begin{IEEEbiography}
[{\includegraphics[trim=4.8cm 6.8cm 4.8cm 5cm, width=1in,height=1.25in,clip,keepaspectratio]{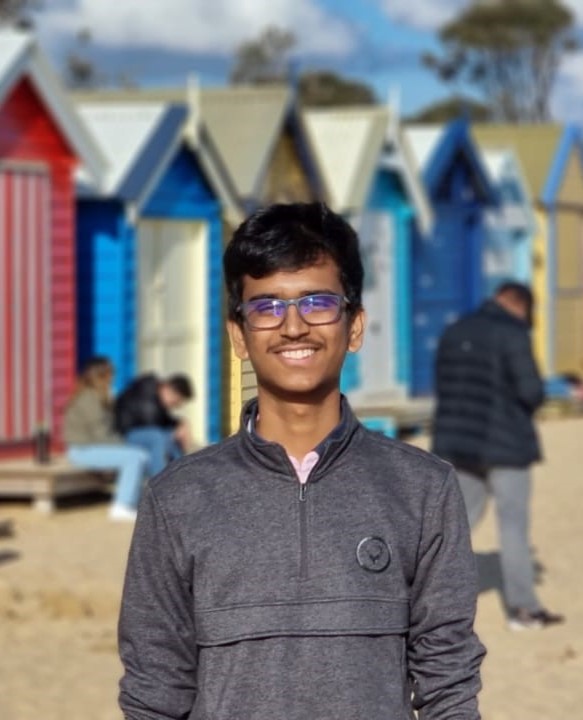}}]{Arjun Raj} is a Research Student at the School of Computing, Australian National University (ANU), supervised by Lei Wang. Previously, he worked as a Research Intern at Active Intelligence Australia Pty Ltd, Perth, where he received the 1st Annual Active Intelligence Research Challenge Award. At Lei's Temporal Intelligence and Motion Extraction (TIME) Lab, Arjun focuses on developing advanced techniques for tracking and detecting high-speed, small objects in sports analysis. His research interests include object detection and tracking, video processing, reusable models, efficient and cost-effective training methods, and promoting sustainability in AI. He aims to merge practical, resource-efficient approaches with cutting-edge AI advancements, contributing to more sustainable and scalable solutions in the field.

\end{IEEEbiography}

\begin{IEEEbiography}
[{\includegraphics[width=1in,height=1.25in,clip,keepaspectratio]{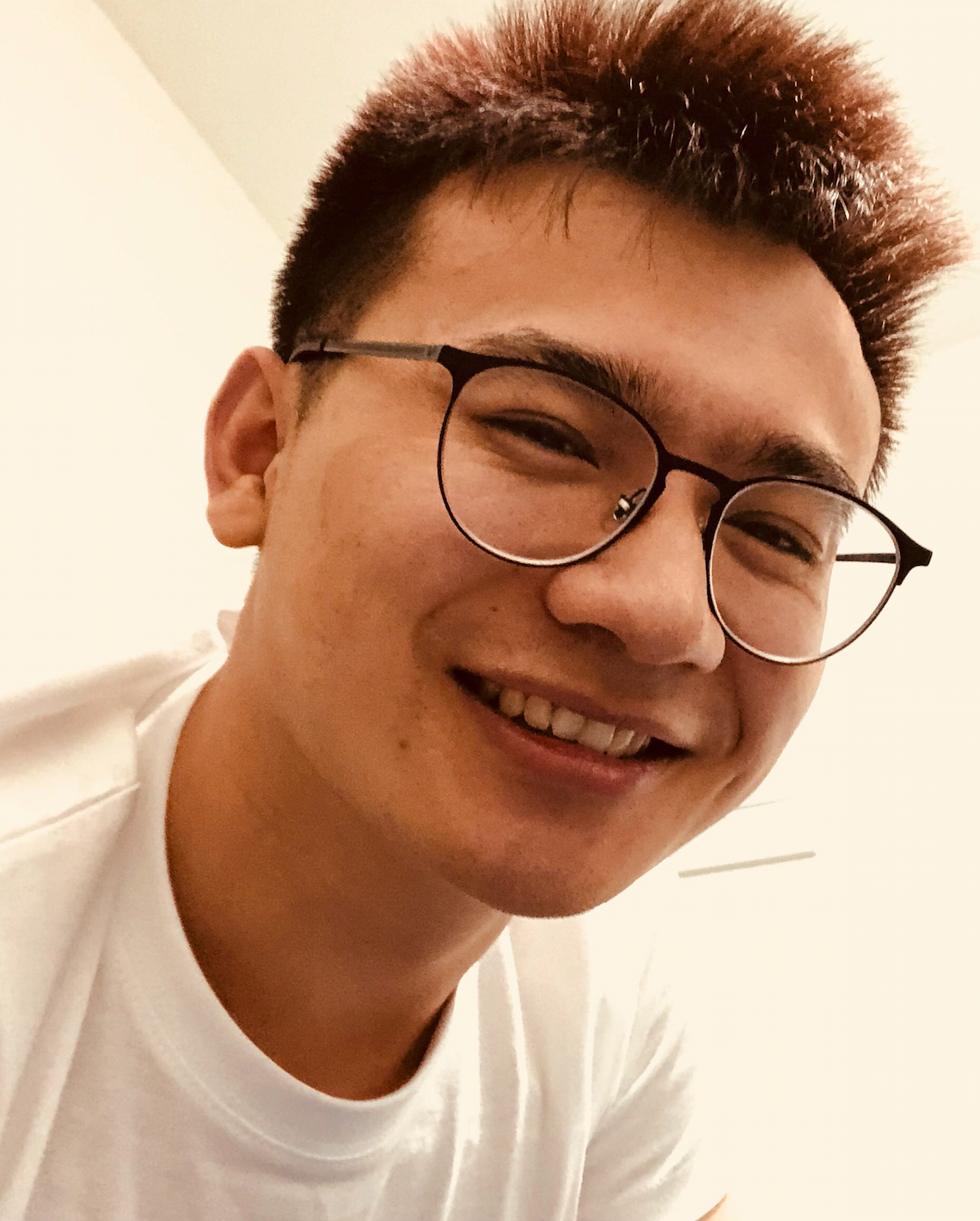}}]{Lei Wang} received his M.E. degree in Software Engineering from The University of Western Australia (UWA), Perth, in 2018, and his Ph.D. in Engineering and Computer Science from the Australian National University (ANU), Canberra, in 2023. He is currently a Research Fellow at the ANU School of Computing, where he leads a dynamic research team of master's and honours students in the Temporal Intelligence and Motion Extraction (TIME) Lab. He is also a Visiting Scientist with the Machine Learning Research Group at Data61/CSIRO (formerly NICTA).
Previously, Lei was a Visiting Researcher at both the Department of Computer Science and Software Engineering at UWA and Data61/CSIRO. Since 2018, he has worked as a full-time Computer Vision Researcher at iCetana Pty Ltd., Perth, and since 2021, he has also served as a Computer Scientist at Active Intelligence Australia Pty Ltd., Perth. Lei has authored numerous first-author papers in prestigious venues, including CVPR, ICCV, ECCV, ACM MM, TPAMI, IJCV, and TIP. He received the Sang Uk Lee Best Student Paper Award at the Asian Conference on Computer Vision (ACCV) 2022.
He currently serves as a Guest Editor for the MDPI open-access journal Electronics (Q2, h-index 83), and as an Area Chair for both the International Conference on Pattern Recognition (ICPR 2024) and ACM Multimedia 2024. His research interests include action recognition, anomaly detection, computer vision, and machine learning. Lei is an active member of IEEE and ACM as a Student Member.
\end{IEEEbiography}

\begin{IEEEbiography}
[{\includegraphics[trim=0.5cm 1.5cm 0.8cm 0cm, width=1in,height=1.25in,clip,keepaspectratio]{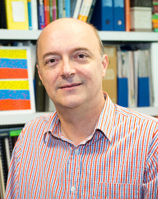}}]{Tom Gedeon}
received the B.Sc. (Hons.) and Ph.D. degrees from the University of Western Australia, Perth, WA, Australia. He holds the Optus Chair in AI and the Director of the Optus Centre for AI, Curtin University, Perth. Before this, he was a Professor of computer science and the former Deputy Dean of the College of Engineering and Computer Science, Australian National University, Canberra, ACT, Australia. He remains an Honorary Professor at ANU. He has over 400 publications. His main research interests include responsive and responsible AI and underpinned by edge computing efficient AI. His focus is on the development of automated systems for information extraction, from eye gaze and physiological data, as well as textual and other data, and for the synthesis of the extracted information into humanly useful information resources, primarily using neural/deep networks and fuzzy logic methods. Prof. Gedeon has run a number of international conferences. He is the former President of the Asia Pacific Neural Network Assembly and the Computing Research and Education Association of Australasia. He is currently a member of the Australian Research Council’s Medical Research Advisory Group. He has been nominated for VC’s awards for postgraduate supervision at three universities. He has been the General Chair of the International Conference on Neural Information Processing (ICONIP) three times. He is an Associate Editor of the IEEE Transactions on Fuzzy Systems and the Neural Networks (INNS/Elsevier).
\end{IEEEbiography}

\end{document}